\documentclass{article}
\usepackage[latin9]{inputenc}
\usepackage{color}
\usepackage{array}
\usepackage{float}
\usepackage{wrapfig}
\usepackage{booktabs}
\usepackage{fancybox}
\usepackage{calc}
\usepackage{url}
\usepackage{bm}
\usepackage{multirow}
\usepackage{amsmath}
\usepackage{amsthm}
\usepackage{graphicx}
\usepackage[authoryear,round]{natbib}
\PassOptionsToPackage{normalem}{ulem}
\usepackage{ulem}
\usepackage[bookmarks=true,bookmarksnumbered=false,bookmarksopen=false,
 breaklinks=false,pdfborder={0 0 1},backref=false,colorlinks=true]
 {hyperref}
\hypersetup{
 citecolor=blue}

\makeatletter

\providecommand{\tabularnewline}{\\}
\floatstyle{ruled}
\newfloat{algorithm}{tbp}{loa}
\providecommand{\algorithmname}{Algorithm}
\floatname{algorithm}{\protect\algorithmname}

\theoremstyle{definition}
\newtheorem*{example*}{\protect\examplename}
\theoremstyle{plain}
\newtheorem{lem}{\protect\lemmaname}
\theoremstyle{remark}
\newtheorem{rem}{\protect\remarkname}

\usepackage{iclr2025_conference}
\usepackage{times}

\usepackage{amsmath,amsfonts,bm}

\def\eqref#1{equation~\ref{#1}}
\def\1{\bm{1}}

\DeclareMathAlphabet{\mathsfit}{\encodingdefault}{\sfdefault}{m}{sl}
\SetMathAlphabet{\mathsfit}{bold}{\encodingdefault}{\sfdefault}{bx}{n}

\usepackage{url}
\usepackage{algorithm}
\usepackage{algorithmic}

\title{Causal Discovery via Bayesian Optimization}

\author{Bao Duong, Sunil Gupta, and Thin Nguyen \\
Applied Artificial Intelligence Institute (A$^2$I$^2$)\\
Deakin University, Geelong, Australia \\
\texttt{\{b.duong,sunil.gupta,thin.nguyen\}@deakin.edu.au}
}

\iclrfinalcopy % Uncomment for camera-ready version, but NOT for submission.

\makeatother

\providecommand{\examplename}{Example}
\providecommand{\lemmaname}{Lemma}
\providecommand{\remarkname}{Remark}

\begin{document}
\maketitle

\global\long\def\ours{\mathbf{DrBO}}%

\global\long\def\pa{\mathrm{pa}}%

\global\long\def\vecdag{\mathtt{Vec2DAG}}%

\global\long\def\indep{\perp\!\!\!\!\perp}%

\vspace{-4mm}

\begin{abstract}
Existing score-based methods for directed acyclic graph (DAG) learning
from observational data struggle to recover the causal graph accurately
and sample-efficiently. To overcome this, in this study, we propose
$\ours$ (\textbf{\uline{D}}AG \textbf{\uline{r}}ecovery via \textbf{\uline{B}}ayesian
\textbf{\uline{O}}ptimization)---a novel DAG learning framework leveraging
Bayesian optimization (BO) to find high-scoring DAGs. We show that,
by sophisticatedly choosing the promising DAGs to explore, we can
find higher-scoring ones much more efficiently. To address the scalability
issues of conventional BO in DAG learning, we replace Gaussian Processes
commonly employed in BO with dropout neural networks, trained in a
continual manner, which allows for (i) flexibly modeling the DAG scores
without overfitting, (ii) incorporation of uncertainty into the estimated
scores, and (iii) scaling with the number of evaluations. As a result,
$\ours$ is computationally efficient and can find the accurate DAG
in fewer trials and less time than existing state-of-the-art methods.
This is demonstrated through an extensive set of empirical evaluations
on many challenging settings with both synthetic and real data. Our
implementation is available at \url{https://github.com/baosws/DrBO}.

\end{abstract}
\vspace{-5mm}

\section{Introduction\label{sec:Introduction}}

\vspace{-2mm}

Learning directed acyclic graphs (DAGs) encoding the underlying causal
relationships, also known as causal discovery, provides invaluable
insights about interventional outcomes and counterfactuals, and thus
significant research effort has been dedicated on this frontier. Our
study focuses on the score-based framework---a major class of causal
discovery methods which casts the DAG learning problem as an optimization
problem, maximizing over the space of DAGs a predefined score function
measuring how a DAG $\mathcal{G}$ fits the observed data $\mathcal{D}$:

\vspace{-6mm}

\begin{equation}
\mathcal{G}^{\ast}=\underset{\mathcal{\mathcal{G}\in\text{DAGs}}}{\arg\max}\ S\left(\mathcal{D},\mathcal{G}\right).\label{eq:score-based}
\end{equation}

\vspace{-4mm}
There are several challenges associated with this formulation. First,
this optimization problem is NP-hard in general \citep{Chickering_96Learning},
due to the combinatorial search domain that scales super-exponentially
with the graph size \citep{Robinson_77Counting} and the acyclicity
condition that is nontrivial to maintain. The second challenge is
computational cost, as the score evaluation can be expensive for complex
models \citep{Zhu_etal_2020Causal,Wang_etal_2021Ordering}, and hence,
methods requiring too many trials will incur a heavy computational
expense.

Bayesian Optimization (BO) has emerged as an effective approach for
expensive black-box optimization thanks to its sample efficiency.\footnote{``Sample'' in ``sample efficiency'' refers to the number of trials.}
Its applicability covers many domains due to the pervasiveness of
optimization tasks in virtually every field. The central idea is that
past evaluation data can reveal potential candidates to evaluate next,
and effectively exploiting them allows us to arrive at better solutions
using fewer evaluations. However, while BO has been employed for \textit{active
causal discovery} \citep{Toth_etal_22Active,Zhang_etal_24Bayesian}
to suggest cost-effective intervention strategies to discover causal
graphs from active experiments, its power is yet to be harnessed for
the problem of \textit{observational causal discovery}, where no active
intervention is available.

We have identified two potential difficulties preventing BO to optimize
Eq.~(\ref{eq:score-based}). The first is the limited scalability
of BO in general, which is usually restricted to only a few hundred
dimensions and thousands of evaluations \citep{Wang_etal_23Recent},
while DAG learning often involves far more trials \citep{Zhu_etal_2020Causal,Wang_etal_2021Ordering,Duong_etal_24Alias}.
The second challenge is how to efficiently optimize the ``acquisition
function'' measuring the potential of DAG solutions in BO. This itself
is a score-based DAG learning problem and is expected to be cheaper
than the original problem as it is repeated many times in the BO pipeline,
thus requiring to be very efficient to be practical.

\textbf{Present study.} In this study, we tackle these obstacles and
bring forth the benefits of BO into causal discovery with the introduction
of $\ours$ (\textbf{\uline{D}}AG \textbf{\uline{r}}ecovery via \textbf{\uline{B}}ayesian
\textbf{\uline{O}}ptimization)---a novel causal discovery algorithm
that leverages BO to find the highest-scoring DAG.\textit{} More
particularly, we solve the aforementioned challenges by several key
design choices. (\textcolor{red}{i}) Inspired by \citet{Yu_etal_2021Dags,Massidda_etal_2023Constraint,Duong_etal_24Alias},
we devise a low-rank DAG representation that relaxes the constrained
optimization in Eq.~(\ref{eq:score-based}) to an \textit{unconstrained
optimization problem }with a dimensionality growing linearly with
the number of nodes (Sec.~\ref{subsec:Search-Space}), enabling the
use of BO with an amenable dimensionality. (\textcolor{red}{ii}) We
replace the Gaussian processes (GPs) in conventional BO approaches
that scale cubically w.r.t. the number of evaluations with \textit{dropout
neural networks} \citep{Srivastava_etal_14Dropout,Guo_etal_21Evolutionary},
which offer expressive uncertainty-aware modeling capabilities, as
well as faster acquisition function calculation and optimization (Sec.~\ref{subsec:Surrogate-Modeling}).
(\textcolor{red}{iii}) Our surrogate model \textit{learns the DAG
score indirectly} via node-wise local scores, allowing us to predict
the DAG scores more accurately with enhanced training information,
compared with learning a direct map from a DAG to its score and not
exploiting the local scores (Sec.~\ref{subsec:Decomposable-Surrogate}).
(\textcolor{red}{iv}) Instead of being retrained every step with all
data, our neural networks are \textit{trained continually}, enabling
our method to scale better with the number of trials (Sec.~\ref{subsec:Continual-Model-Training}).
Compared to existing optimization strategies in causal discovery,
like gradient-based methods, which do not attempt to exploit past
exploration data to prioritize visiting the most promising DAGs, our
method makes more informed decisions about which DAGs to investigate
next, leading to fewer unnecessary trials to reach high-scoring DAGs.

\textbf{Contributions.} The main contributions of our study are summarized
as follows:

\vspace{-2.5mm}

\begin{enumerate}
\item To facilitate sample-efficient causal discovery, we propose $\ours$---a
causal discovery method employing BO to optimize for the DAG score.
$\ours$ is specifically designed for causal discovery, aiming to
be not only accurate, but also computationally manageable. \textit{To
our knowledge, this is the first score-based causal discovery method
based on BO for purely observational data.}
\item We demonstrate the effectiveness of our method on a comprehensive
set of experiments, showing that $\ours$ can consistently surpass
state-of-the-art baselines on various conditions, with fewer evaluations
and less time, signifying the sample efficiency of BO in our design.
In addition, extensive ablation studies verify the significance of
our design choices.
\end{enumerate}

\vspace{-5mm}

\section{Related Work\label{sec:Related-Work}}

\vspace{-4mm}

Causal discovery methods can be broadly categorized into two major
classes, namely constraint-based and score-based methods.\textbf{
}The former category \citep{Spirtes_etal_00Causation,Colombo_etal_12Learning}
involves performing a series of hypothesis tests to recover the undirected
skeleton of the causal graph, before orienting the edges using graphical
rules. On the other hand, score-based causal discovery recasts the
problem as a combinatorial optimization task. Classical methods \citep{Chickering_02Optimal,Ramsey_17Million,Ramsey_15Scaling}
greedily traverse the DAG space by adding and removing edges one-at-a-time
to maintain acyclicity. Recent advances include relaxing the combinatorial
optimization to a continuous optimization problem \citep{Zheng_etal_18DAGs,Yu_etal_2019Dag,Zheng_etal_20Learning,Yu_etal_2021Dags,Zhang_etal_2022Truncated,Bello_etal_22dagma,Annadani_etal_2023Bayesdag}.
In addition, methods based on Reinforcement learning (RL) \citep{Zhu_etal_2020Causal,Wang_etal_2021Ordering,Yang_etal_2023Reinforcement,Duong_etal_24Alias}
have recently emerged as competitive search strategies. We also
acknowledge \textit{interventional causal discovery} studies \citep{Hauser_Buhlmann_12Characterization,Brouillard_etal_20Differentiable,Lippe_etal_22Efficient}
where interventional data is exploited to help identify the causal
DAG, however, here we focus on the more challenging setting where
no intervention is available.

Furthermore, \textit{Bayesian causal discovery} studies are an intriguing
direction \citep{Deleu_etal_22Bayesian,Tran_etal_23Differentiable,Annadani_etal_2023Bayesdag},
where the aim is to infer the posterior distribution over causal graphs
given observed data, in order to quantify uncertainty in \textit{DAG
estimates}. Meanwhile, our use of BO involves quantifying uncertainty
in \textit{DAG scores}, so despite sharing the term ``Bayesian'',
these studies are distant from us. Moreover, while Bayesian optimization
has been employed for causal discovery in the \textit{active} setting
\citep{Toth_etal_22Active,Zhang_etal_24Bayesian}, these methods utilize
BO to suggest optimal interventions to quickly recover the causal
DAG, necessitating the ability to perform active experiments. Meanwhile,
our study utilizes BO to optimize for the score function that can
be calculated purely from observational data. In addition, \textit{causal
Bayesian optimization} \citep{Aglietti_etal_20Causal,Aglietti_etal_21Dynamic},
which deals with optimizing a variable of interest that is part of
a causal system\textcolor{red}{{} }with \textit{known DAG} via a series
of interventions suggested by BO, is also a different problem from
ours, which focuses on finding the unknown DAG instead.

\section{Background\label{sec:Background}}

\vspace{-3mm}

\textbf{Notations.} In this paper, unless specifically indicated,
normal lowercase letters indicate scalars (e.g., $x$, $y$) or functions
(e.g., $f$, $g$), while bold lowercase letters represent vectors
(e.g., $\mathbf{x}$, $\mathbf{y}$), and bold uppercase letters denote
matrices (e.g., $\mathbf{X}$). Meanwhile, subscripts and bracketed
superscripts index dimensions and samples, respectively, e.g., $x_{i}^{\left(j\right)}$
denotes the $i$-th dimension of the $j$-th sample.

\vspace{-3mm}

\subsection{Bayesian Optimization\label{subsec:Bayesian-Optimization}}

\vspace{-2mm}

We provide here only the details necessary to understand our contributions.
For a more comprehensive review of BO, see \citet{Wang_etal_23Recent},
for example. Consider a maximization of a function $f$ that is expensive
to evaluate: $\mathbf{x}^{\ast}=\arg\max_{\mathbf{x}\in\mathcal{X}}f\left(\mathbf{x}\right)$.
BO is a class of sequential optimization methods, which iteratively
(i) proposes potential candidate(s) to evaluate based on an ``acquisition
function'', then (ii) evaluates said candidate(s), and (iii) updates
its statistical model with the newly acquired observations. More
specifically, BO defines a probabilistic ``surrogate'' model over
the distribution $f$ given observed data $\mathbf{X},\mathbf{y}$,
i.e., $P\left(f\mid\mathbf{X},\mathbf{y}\right)$, to define the acquisition
function.

\textbf{Surrogate model.} To model $P\left(f\mid\mathbf{X},\mathbf{y}\right)$,
Gaussian process (GP) is the standard in BO since it offers a closed-form
solution \citep{Rasmussen_03Gaussian}. Assuming we have observed
a dataset of $n$ evaluations $\mathbf{X}=\left[\mathbf{x}^{\left(1\right)},\ldots,\mathbf{x}^{\left(n\right)}\right]^{\top}$
and $\mathbf{y}=\left[y^{\left(1\right)},\ldots,y^{\left(n\right)}\right]^{\top}$,
where $\mathbf{x}\sim\mathcal{X}=\mathbb{R}^{d}$ and $y:=f\left(\mathbf{x}\right)$,
GP assumes that $\mathbf{y}$ follows a multivariate Gaussian distribution
governed by a mean function $\mu:\mathcal{X}\rightarrow\mathbb{R}$
and positive-definite covariance function $\kappa$$:\mathcal{X}\times\mathcal{X}\rightarrow\mathbb{R}$:
$\mathbf{y}\sim\mathcal{N}\left(\mu\left(\mathbf{x}\right),\mathbf{K}_{\mathbf{X,X}}\right)$
where $\mathbf{K}_{\mathbf{X,X}}:=\left[\kappa\left(\mathbf{x}^{\left(i\right)},\mathbf{x}^{\left(j\right)}\right)\right]_{i,j=1,\ldots,n}$
is the $n\times n$ covariance matrix. Using Bayes' theorem, the posterior
of the function value at a new location $\mathbf{x}$ is given analytically
as: $P\left(y\mid\mathbf{x},\mathbf{X},\mathbf{y}\right)=\mathcal{N}\left(\mu\left(\mathbf{x}\right),\sigma^{2}\left(\mathbf{x}\right)\right)$
where $\mu\left(\mathbf{x}\right):=\mathbf{k}_{\mathbf{x,X}}\mathbf{K}_{\mathbf{X,X}}^{-1}\mathbf{y}$
and $\sigma^{2}\left(\mathbf{x}\right):=\kappa\left(\mathbf{x},\mathbf{x}\right)-\mathbf{k}_{\mathbf{X,x}}^{\top}\mathbf{K}_{\mathbf{X},\mathbf{X}}^{-1}\mathbf{k}_{\mathbf{X},\mathbf{x}}$.
GPs scale poorly with $n$ due to the need to invert $\mathbf{K}_{\mathbf{X},\mathbf{X}}$,
so alternative statistical models like random forest \citep{Hutter_etal_11Sequential},
Bayesian linear regression \citep{Snoek_etal_15Scalable}, and Bayesian
neural network \citep{Springenberg_etal_16Bayesian} have been employed
as more scalable surrogate models.

\textbf{Acquisition functions} (AFs) in BO judge how promising an
arbitrary candidate $\mathbf{x}$ is based on the posterior inferred
by the surrogate model, to make a more informed candidate suggestion.
As an example, upper confidence bound (UCB) is a common choice, designed
to minimize regret in the multi-armed bandit literature \citep{Srinivas_etal_09Gaussian}
and given by $\mathrm{AF}\left(\mathbf{x}\right):=\mu\left(\mathbf{x}\right)+\beta\sigma\left(\mathbf{x}\right)$,
where $\beta>0$ is a hyperparameter controlling the exploitation-exploration
trade-off. Another promising acquisition is Thompson sampling \citep[TS, ][]{Thompson_33Likelihood},
which is a stochastic function that uses a random draw from the posterior
as the potential indicator: $\mathrm{AF}\left(\mathbf{x}\right)\sim P\left(y\mid\mathbf{x},\mathbf{X},\mathbf{y}\right)$.

\vspace{-3mm}

\subsection{Structural Causal Model}

\vspace{-2mm}

Let $\mathbf{x}\in\mathbb{R}^{d}$ be the random vector capturing
the system of interest, and $\mathcal{D}=\left\{ \mathbf{x}^{\left(k\right)}\right\} _{k=1}^{n}$
denote an i.i.d. dataset of $n$ samples from $P\left(\mathbf{x}\right)$.
The structural causal model \citep[SCM, ][]{Pearl_2000Models,Pearl_09Causality}
among said variables can be described by (i) a DAG $\mathcal{G}=\left(\mathcal{V},\mathcal{E}\right)$
where each node $i\in\mathcal{V}=\left\{ 1,\ldots,d\right\} $ corresponds
to a random variable $x_{i}$, and each edge $\left(j\rightarrow i\right)\in\mathcal{E}\subset\mathcal{V}\times\mathcal{V}$
implies that $x_{j}$ is a direct cause of $x_{i}$, (ii) a set of
functions $\left\{ f_{i}\right\} _{i=1}^{n}$ dictating the causal
mechanisms, and (iii) a noise distribution $P\left(\bm{\varepsilon}\right)$.
Together, these components define a generative process $x_{i}:=f_{i}\left(\mathbf{x}_{\mathrm{pa}_{i}^{\mathcal{G}}},\varepsilon_{i}\right),\forall i=1,\ldots,d$,
where $\pa_{i}^{\mathcal{G}}=\left\{ j\in\mathcal{V}\mid\left(j\rightarrow i\right)\in\mathcal{E}\right\} $
is the set of direct causes (a.k.a. structural parents) of node $i$
in $\mathcal{G}$, and $\bm{{\bf \varepsilon}}\sim P\left(\bm{\varepsilon}\right)$
is the noise vector. Then, the observational causal discovery problem
is concerned about recovering the DAG $\mathcal{G}$ from the observational
dataset $\mathcal{D}$. In addition, following \citep{Zhu_etal_2020Causal,Wang_etal_2021Ordering,Yang_etal_2023Reinforcement,Yang_etal_2023Causal,Duong_etal_24Alias},
we also assume: (i)\textit{ causal sufficiency}: there is no unobserved
confounders among the variables;\textit{ }(ii) \textit{causal minimality}:
there is no function $f_{i}$ that is constant to any of its argument
\citep{Peters_etal_2014Causal}; and\textit{ }(iii)\textit{ identifiable
causal model}s: this means $\mathcal{G}$ is the unique causal graph
that can induce $P\left(\mathbf{x}\right)$, and thus it is possible
to be recovered. For instance, while general linear-Gaussian models
are known to be unidentifiable \citep{Spirtes_etal_00Causation},
examples for identifiable causal models include linear-Gaussian models
with the equal-variance assumption \citep{Peters_etal_2014Causal}
and nonlinear additive noise models (ANMs) in general \citep{Hoyer_etal_2008Nonlinear}.
Our experiments will adopt these identifiable models.

\vspace{-3mm}

\subsection{Score-based Causal Discovery}

\vspace{-2mm}

A critical component of score-based causal discovery is the proper
specification of a scoring function. With the proper scoring function,
the optimization of the score is equivalent to reaching to ground
truth DAG. Consistent scoring functions \citep{Chickering_02Optimal}
are known to satisfy this requirement. In the main text, we demonstrate
our method with the Bayesian Information Criterion \citep[BIC, ][]{Schwarz_1978Estimating},
which is a consistent score as shown by \citet{Haughton_88Choice}
and is widely considered in numerous existing studies \citep{Chickering_02Optimal,Zhu_etal_2020Causal,Wang_etal_2021Ordering,Yang_etal_2023Reinforcement,Yang_etal_2023Causal,Duong_etal_24Alias}.

More formally, let $\theta:=\left\{ \left\{ f_{i}\right\} _{i=1}^{d},P\left(\bm{\varepsilon}\right)\right\} $
be the parameters of an SCM, then the BIC score is given by $S_{\text{BIC}}\left(\mathcal{D},\mathcal{G}\right):=2\ln p\left(\mathcal{D}\mid\hat{\theta},\mathcal{G}\right)-\left|\mathcal{G}\right|\ln n$,
where $\hat{\theta}:=\arg\max_{\theta}p\left(\mathcal{D}\mid\theta,\mathcal{G}\right)$
is the maximum-likelihood estimator of the causal model parameters,
$n$ is the sample size of $\mathcal{D}$, and $\left|\mathcal{G}\right|$
denotes the number of edges in $\mathcal{G}$. The generality of
BIC allows for its adaptation to numerous causal models. In the main
paper, we showcase our method with the BIC defined for the popular
additive noise model \citep[ANM, ][]{Hoyer_etal_2008Nonlinear}: $x_{i}:=f_{i}\left(\mathbf{x}_{\pa_{i}}\right)+\varepsilon_{i}$,
where $\varepsilon_{i}\sim\mathcal{N}\left(0,\sigma_{i}^{2}\right)$.
In the general form, where the noise variances can be non-equal, the
BIC for ANM is specified as follows:

\vspace{-5mm}

\begin{equation}
S_{\text{BIC-NV}}\left(\mathcal{D},\mathcal{G}\right):=-n\sum_{i=1}^{d}\ln\text{MSE}_{i}\left(\pa_{i}^{\mathcal{G}}\right)-\left|\mathcal{G}\right|\ln n,\label{eq:bic-nv}
\end{equation}

\vspace{-4mm}

where NV stands for ``non-equal variance'' and $\text{MSE}_{i}\left(\pa_{i}^{\mathcal{G}}\right):=\frac{1}{n}\sum_{j=1}^{n}\left(x_{i}^{\left(j\right)}-\hat{f}_{i}\left(\mathbf{x}_{\pa_{i}^{\mathcal{G}}}^{\left(j\right)}\right)\right)^{2}$
is the mean squared error after regressing $x_{i}$ on $\mathbf{x}_{\pa_{i}^{\mathcal{G}}}$.
In addition, if we further assume that the noise variables have \uline{e}qual
\uline{v}ariances \citep{Buhlmann_etall_14Cam} then the BIC yields:

\vspace{-3mm}
\begin{equation}
S_{\text{BIC-EV}}\left(\mathcal{D},\mathcal{G}\right):=-nd\ln\frac{\sum_{i=1}^{d}\text{MSE}_{i}\left(\pa_{i}^{\mathcal{G}}\right)}{d}-\left|\mathcal{G}\right|\ln n.\label{eq:bic-ev}
\end{equation}

Eqs.~(\ref{eq:bic-nv}) and (\ref{eq:bic-ev}) are common in prior
studies \citep{Zhu_etal_2020Causal,Wang_etal_2021Ordering,Yang_etal_2023Reinforcement,Yang_etal_2023Causal,Duong_etal_24Alias},
and we also provide their derivation in Appendix~\ref{subsec:Derivation-of-BIC}.

\subsection{Parametrized DAG generation}

\vspace{-2mm}

For effective acquisition function optimization, we find it crucial
for our method to be able to quickly generate candidate DAGs within
specific regions determined by a low-dimensional search space. This
would help narrow down the regions of interest and improve the quality
of suggested candidates. A potential approach towards this end is
autoregressive DAG generation techniques \citep{Wang_etal_2021Ordering,Deleu_etal_22Bayesian,Yang_etal_2023Reinforcement,Deleu_etal_24Joint},
which break down the DAG generation into multiple sequential steps.
However, each step of the generation process is computationally involved
due to their autoregressive nature \citep{Duong_etal_24Alias}, accumulating
into a significant DAG generation cost. Recently, constraint-free
DAG representations with a quadratic complexity are proposed in \citep{Yu_etal_2021Dags,Massidda_etal_2023Constraint,Duong_etal_24Alias},
which introduce different maps from an unconstrained real-valued representation
to the space of DAGs. For example, the $\vecdag$ operator in \citep{Duong_etal_24Alias}
takes as input a continuous ``node potential'' vector $\mathbf{p}\in\mathbb{R}^{d}$
and strictly upper-triangular ``edge potential'' matrix $\mathbf{E}\in\mathbb{R}^{d\times d}$
to deterministically create a DAG: $\vecdag\left(\mathbf{p},\mathbf{E}\right):=H\left(\mathrm{grad}\left(\mathbf{p}\right)\right)\odot H\left(\mathbf{E}+\mathbf{E}^{\top}\right)$,
where $H\left(x\right):=\begin{cases}
1, & \text{if }x>0,\\
0, & \text{otherwise}
\end{cases}$ is the entry-wise Heaviside step function, $\odot$ is the Hadamard
product, and $\mathrm{grad}\left(\mathbf{p}\right)_{ij}:=p_{j}-p_{i}$
is the gradient flow operator \citep{Lim_20Hodge}.

\begin{example*}
Consider a system of 3 nodes with potentials $\mathbf{p}=\left[-1,3,2\right]$
and $\mathbf{E}=\left[\begin{array}{ccc}
0 & 2 & -4\\
0 & 0 & 7\\
0 & 0 & 0
\end{array}\right]$. Then, $\vecdag\left(\mathbf{p},\mathbf{E}\right)=H\left(\left[\begin{array}{ccc}
0 & 4 & 3\\
-4 & 0 & -1\\
-3 & 1 & 0
\end{array}\right]\right)\odot H\left(\left[\begin{array}{ccc}
0 & 2 & -4\\
2 & 0 & 7\\
-4 & 7 & 0
\end{array}\right]\right)=\left[\begin{array}{ccc}
0 & 1 & 0\\
0 & 0 & 0\\
0 & 1 & 0
\end{array}\right]$. This is the adjacency of the DAG $1\rightarrow2\leftarrow3$, where
the edge directions and connectivities are determined by the first
and second terms of $\vecdag$, respectively.
\end{example*}
Using this approach, it is simple to sample candidate DAGs whose representations
are in the neighborhood around some values $\mathbf{p}$ and $\mathbf{E}$
of interest.

\section{$\protect\ours$: DAG Recovery via Bayesian Optimization\label{sec:Method}}

\begin{algorithm}[t]
\caption{The $\protect\ours$ method for causal discovery.\label{alg:ours}}

\begin{algorithmic}[1]

\REQUIRE{Dataset $\mathcal{D}=\left\{ \mathbf{x}^{\left(j\right)}\in\mathbb{R}^{d}\right\} _{j=1}^{n}$
of $d$ nodes and $n$ observations, score function $S\left(\mathcal{D},\cdot\right)$,
DAG rank $k$, batch size $B$, no. of preliminary candidates $C$,
and total no. of evaluations $T$.}

\ENSURE{A DAG $\hat{\mathcal{G}}$ that maximizes $S\left(\mathcal{D},\mathcal{G}\right)$.}

\STATE Initialize empty experience $\mathcal{H}:=\emptyset$ and
node-wise dropout neural nets: $\left\{ \mathrm{DropoutNN}_{i}\right\} _{i=1}^{d}$.

\WHILE{$\left|\mathcal{H}\right|<T$}

\STATE Generate random DAGs: $\left\{ \mathcal{G}^{\left(j\right)}:=\tau\left(\mathbf{z}^{\left(j\right)}\right)\right\} _{j=1}^{C}$
where $\mathbf{z}\in\left[-1,1\right]^{d\left(1+k\right)}$.\hfill$\triangleright$
Sec.~\ref{subsec:Search-Space}.

\STATE Sample local scores: $\left\{ \left\{ l_{i}^{\left(j\right)}\sim\mathrm{DropoutNN}_{i}\left(\pa_{i}^{\mathcal{G}^{\left(j\right)}}\right)\right\} _{i=1}^{d}\right\} _{j=1}^{C}$.\hfill$\triangleright$
Sec.~\ref{subsec:Surrogate-Modeling}.

\STATE Combine local scores: $\left\{ \mathrm{AF}^{\left(j\right)}:=\mathrm{Combine}\left(l_{1}^{\left(j\right)},\ldots,l_{d}^{\left(j\right)}\right)\right\} _{j=1}^{C}$.\hfill$\triangleright$
Sec.~\ref{subsec:Decomposable-Surrogate}.

\STATE Select top $B$ candidates with highest AF values: $j_{1},\ldots,j_{B}:=\underset{j=1,\ldots,C}{\text{argtop}_{B}}\ \mathrm{AF}^{\left(j\right)}$.\hfill$\triangleright$
Sec.~\ref{subsec:Acquisition-Function-Optimizatio}.

\STATE Evaluate these candidates and update experience: $\mathcal{H}:=\mathcal{H}\cup\left\{ \left(\mathcal{G}^{\left(j\right)},S\left(\mathcal{D},\mathcal{G}^{\left(j\right)}\right)\right)\right\} _{j=j_{1},\ldots,j_{B}}$.

\STATE Update the neural nets on new $\mathcal{H}$.\hfill$\triangleright$
Sec.~\ref{subsec:Continual-Model-Training}.

\ENDWHILE

\STATE Get highest-scoring DAG so far: $\hat{\mathcal{G}}:=\arg\max_{\mathcal{G}\in\mathcal{H}}S\left(\mathcal{D},\mathcal{G}\right)$.

\STATE Prune $\hat{\mathcal{G}}$ if needed.\hfill$\triangleright$
Sec.~\ref{subsec:Pruning}.

\end{algorithmic}
\end{algorithm}

\vspace{-2mm}

An overview of our framework is illustrated in Algorithm~\ref{alg:ours}.
In the following, we describe the proposed $\ours$ algorithm step-by-step.

\vspace{-2mm}

\subsection{Search Space\label{subsec:Search-Space}}

\vspace{-3mm}

To effectively utilize BO, the search space should be unconstrained.
Thus, following \citet{Yu_etal_2021Dags,Massidda_etal_2023Constraint,Duong_etal_24Alias},
we transform the constrained combinatorial optimization problem in
Eq.~(\ref{eq:score-based}) to an unconstrained optimization task.
However, the dimensionality of $\mathcal{O}\left(d^{2}\right)$ of
their search spaces can still be reduced to mitigate the effect of
the curse of dimensionality, facilitating easier acquisition function
optimization. For this purpose, similarly to \citet{Fang_etal_23Low},
we assume that the edge potential matrix in $\vecdag$ \citep{Duong_etal_24Alias}
is low-rank and thus consider an adaptation that offers a search space
that grows linearly with the number of nodes. Specifically, each node
$i$ is now associated with a low-dimensional embedding vector $r_{i}\in\mathbb{R}^{k}$
with $k\ll d$, and two nodes $i$ and $j$ are connected if and only
if $\left\langle r_{i},r_{j}\right\rangle >0$. The total dimensionality
of this search space is thus only $d\cdot\left(1+k\right)$. More
formally, given a node potential $\mathbf{p}\in\mathbb{R}^{d}$ and
an embedding matrix $\mathbf{R}\in\mathbb{R}^{d\times k}$, we define
the following map

\vspace{-3mm}
\begin{equation}
\tau\left(\mathbf{p},\mathbf{R}\right):=H\left(\mathrm{grad}\left(\mathbf{p}\right)\right)\odot H\left(\mathbf{R}\cdot\mathbf{R}^{\top}\right).\label{eq:vec2dag-lr}
\end{equation}

\vspace{-2mm}

The following Lemma ensures the acyclicity of the DAG corresponding
to $\tau\left(\mathbf{p},\mathbf{R}\right)$.
\begin{lem}
\label{lem:vec2dag-lr}For all $d,k\in\mathbb{N}^{+}$, $\mathbf{p}\in\mathbb{R}^{d}$
and $\mathbf{R}\in\mathbb{R}^{d\times k}$, let $\tau:\mathbb{R}^{d}\times\mathbb{R}^{d\times k}\rightarrow\left\{ 0,1\right\} ^{d\times d}$
be defined as in Eq.~(\ref{eq:vec2dag-lr}). Then, $\tau\left(\mathbf{p},\mathbf{R}\right)$
represents a binary adjacency matrix of a DAG.
\end{lem}
The proof can be found in Appendix~\ref{subsec:proof-vec2dag-lr}.
In addition, like $\vecdag$, our variation also exhibits the following
scale-invariance property.
\begin{lem}
\label{lem:scale-invariance}For all $d,k\in\mathbb{N}^{+}$, $\mathbf{p}\in\mathbb{R}^{d}$
and $\mathbf{R}\in\mathbb{R}^{d\times k}$, let $\tau:\mathbb{R}^{d}\times\mathbb{R}^{d\times k}\rightarrow\left\{ 0,1\right\} ^{d\times d}$
be defined as in Eq.~(\ref{eq:vec2dag-lr}). Then, for all $\alpha>0$,
$\tau\left(\mathbf{p},\mathbf{R}\right)=\tau\left(\alpha\mathbf{p},\alpha\mathbf{R}\right)$.
\end{lem}
The proof is provided in Appendix~\ref{Proof-of-scale-invariance}.
This insight allows us to restrict the search domain to a fixed range
(e.g., $\left[-1,1\right]$) for numerical stability. For brevity,
the remaining parts of this manuscript will use vector $\mathbf{z}$
of $d\cdot\left(1+k\right)$ dimensions as the concatenation of $\mathbf{p}$
and the flattened $\mathbf{R}$, and we adopt the notation $\tau\left(\mathbf{z}\right)\equiv\tau\left(\mathbf{p},\mathbf{R}\right)$.
In short, we can now translate the original optimization problem in
Eq.~(\ref{eq:score-based}) to the following unconstrained optimization
problem:

\vspace{-2mm}

\noindent\Ovalbox{\begin{minipage}[t]{1\columnwidth - 2\fboxsep - 1.6pt}%
\begin{equation}
\mathbf{z^{\ast}}:=\underset{\mathbf{z}\in\mathbb{R}^{d(1+k)}}{\arg\max}S\left(\mathcal{D},\tau\left(\mathbf{z}\right)\right).\label{eq:relaxed}
\end{equation}
\end{minipage}}

\vspace{-1mm}

Obviously, if $k\geq d$ then $\mathbf{R}\cdot\mathbf{R}^{\top}$
is full-rank, so $\tau$ can represent any DAG possible \citep[Theorem 1,][]{Duong_etal_24Alias},
and thus $\mathcal{G}^{\ast}:=\tau\left(\mathbf{z}^{\ast}\right)$
is a maximizer of Eq.~(\ref{eq:score-based}) for any maximizer $\mathbf{z}^{\ast}$
of Eq.~(\ref{eq:relaxed}). While this may not hold for $k<d$, our
empirical evaluations reveal that this representation suffices even
for very complex graphs. Further, we find that lower ranks typically
result in better sample-efficiency than higher ranks (Figure~\ref{fig:Ablating}(a))
and provide an explanation in Appendix~\ref{subsec:Why-low-rank-DAG}.

\subsection{Acquisition Function Optimization\label{subsec:Acquisition-Function-Optimizatio}}

\vspace{-2mm}

To perform each step of the BO pipeline (Sec.~\ref{subsec:Bayesian-Optimization}),
we use the acquisition function obtained so far to select a batch
of $B$ most promising candidates to evaluate, also known as Batch
BO \citep{Joy_etal_20Batch}. This is done by first generating $C\geq B$
preliminary candidates $\left\{ \mathbf{z}^{\left(j\right)}\right\} _{j=1}^{C}$
from a hypercube centered at the current best solution $\mathbf{z^{\ast}}$
(see Appendix~\ref{sec:Latin-Hypercube-Design} for more details).
Subsequently, we evaluate the acquisition function of the DAGs induced
by these candidates,\footnote{Our AFs do not take as input $\mathbf{z}$, but $\tau\left(\mathbf{z}\right)$
instead, as many $\mathbf{z}$'s can produce the same DAG.} and choose the top $B$ candidates based on the acquisition function
values. The quality of the candidates batch strongly depends on $C$,
i.e., if we can evaluate the acquisition function of a lot of candidates,
then the top $B$ candidates are likely to have higher values. However,
this also increases computational cost, so our acquisition function
must scale well with the number of candidates $C$ to mitigate this
overhead, which leads us to the next point.

\subsection{Surrogate Modeling with Dropout Networks\label{subsec:Surrogate-Modeling}}

\vspace{-2mm}

To overcome the scalability issues of standard BO due to the use
of GPs as discussed earlier, we instead pursue neural networks, which
are well-known for their scalability and flexibility \citep{Snoek_etal_15Scalable}.
Our networks must be able to model uncertainty to help the optimizer
prioritize evaluating uncertain but promising candidates. Towards
this end, we employ dropout activations \citep{Srivastava_etal_14Dropout},
whose original purpose was to reduce overfitting in training neural
networks, and were later found to be also useful as an approximate
Bayesian inference method \citep{Gal_Ghahramani_16Dropout}, and thus
have been successfully applied to BO \citep{Guo_etal_21Evolutionary}.
We provide a detailed discussion on this choice compared with other
models in Appendix~\ref{sec:Discussion-on-scalable-surrogate}.

Specifically, we devise a single-layer neural network with dropout
activation as follows. Let $p\in\left(0,1\right)$ be the dropout
rate, $d$ denotes the dimensionality of the input, $h$ is the number
of hidden units, $\mathbf{W}_{1}\in\mathbb{R}^{d\times h}$ and $\mathbf{W}_{2}\in\mathbb{R}^{h\times1}$
are weight matrices, $\mathbf{b}_{1}\in\mathbb{R}^{h}$ and $b_{2}\in\mathbb{R}$
are biases. Our dropout networks are then defined as:

\vspace{-4mm}

\[
\mathrm{DropoutNN}\left(\mathbf{x}\right):=\mathbf{W}_{2}^{\top}\left(\mathrm{BatchNorm}\left(\mathrm{ReLU}\left(\frac{1}{1-p}\left(\left(1-\mathbf{m}\right)\circ\left(\mathbf{W}_{1}^{\top}\mathbf{x}+\mathbf{b}_{1}\right)\right)\right)\right)\right)+b_{2},
\]

\vspace{-3mm}

where we also follow common practice to employ Batch Normalization
\citep{Ioffe_Szegedy_15Batch} for improved training efficiency, and
$\mathbf{m}\sim\mathrm{Bernoulli}\left(p\right)^{h}$ is drawn for
every invocation of $\mathrm{DropoutNN}\left(\mathbf{x}\right)$ in
both train and test modes. By training this model on observed data
$\mathbf{X}$ and $\mathbf{y}$ with the square loss, performing a
stochastic forward pass $y\sim\mathrm{DropoutNN}\left(\mathbf{x}\right)$
can be interpreted as drawing from an approximate posterior $y\sim q\left(y\mid\mathbf{x},\mathbf{X},\mathbf{y}\right)$
\citep{Gal_Ghahramani_16Dropout}. Using Thompson sampling, we do
not need to characterize the whole posterior and this mere sample
suffices for acquisition function optimization \citep{Russo_VanRoy_14Learning,Eriksson_etal_19Scalable}.

\vspace{-1mm}

\subsection{Indirect Surrogate Modeling\label{subsec:Decomposable-Surrogate}}

\vspace{-1mm}

A naïve surrogate modeling approach for DAG learning is modeling a
direct map from a DAG to its score. However, DAG scores can typically
be decomposed into independent node-wise components, which can be
further exploited for better modeling. For example, the BIC scores
in Eqs.~(\ref{eq:bic-nv}) and (\ref{eq:bic-ev}) involve the local
components $\left\{ \mathrm{MSE}_{i}\left(\pa_{i}^{\mathcal{G}}\right)\right\} _{i=1}^{d}$
observable after each DAG score invocation. To exploit these information
to the fullest, we propose to learn local surrogate models predicting
the node-wise scores, then combine them using the rule in Eq.~(\ref{eq:bic-nv})
or (\ref{eq:bic-ev}), depending on the situation.

Particularly, for each node $i$, we use the evaluation data $\left\{ \left(\pa_{i}^{\mathcal{G}^{\left(j\right)}},\mathrm{MSE}_{i}\left(\pa_{i}^{\mathcal{G}^{\left(j\right)}}\right)\right)\right\} _{j=1}$,
where $\mathcal{G}^{\left(j\right)}$ is the $j$-th evaluated DAG,
to train a dropout network $\mathrm{DropoutNN}_{i}$ predicting $\ln\mathrm{MSE}_{i}$
from $\mathrm{pa}_{i}^{\mathcal{G}}$, which is represented by a binary
vector of $d$ dimensions. The models are independent among all nodes
instead of being shared to avoid spurious correlations. To summarize,
for a dataset of $d$ nodes, we jointly train $d$ local surrogate
models $\left\{ \mathrm{DropoutNN}_{i}\right\} _{i=1}^{d}$. To sample
the DAG score of a graph $\mathcal{G}$, we first sample each local
score $l_{i}\sim\mathrm{DropoutNN}_{i}\left(\pa_{i}^{\mathcal{G}}\right)$,
then combine them with $\mathrm{AF}\left(\mathcal{G}\right):=\mathrm{Combine}_{\mathrm{BIC-NV}}\left(l_{1},\ldots,l_{d}\right):=-n\sum_{i=1}^{d}l_{i}-\left|\mathcal{G}\right|\ln n$,
if the non-equal variance BIC score is being considered, or $\mathrm{AF}\left(\mathcal{G}\right):=\mathrm{Combine}_{\mathrm{BIC-EV}}\left(l_{1},\ldots,l_{d}\right):=-nd\ln\frac{\sum_{i=1}^{d}e^{l_{i}}}{d}-\left|\mathcal{G}\right|\ln n$
for the case of equal variance BIC score, which resemble Eqs.~(\ref{eq:bic-nv})
and (\ref{eq:bic-ev}), respectively.

\vspace{-1mm}

\subsection{Continual Model Training\label{subsec:Continual-Model-Training}}

\vspace{-1mm}

Upon acquiring a batch of $B$ of new evaluations, the neural networks
require retraining to update their weights. To avoid retraining with
all data so far, which scales at least quadratically with the number
of evaluations, we adopt a continual training approach \citep{Wang_etal_24comprehensive}
that only updates the models with the new data and a small portion
of past data to mitigate forgetting. Specifically, we perform $n_{\text{grads}}$
gradient steps within each BO iteration, each of which is calculated
on the $B$ new datapoints and a replay buffer of $n_{\text{replay}}$
past observations using Reservoir Sampling \citep{Vitter_85Random}.

\vspace{-1mm}

\subsection{Finalizing the Result\label{subsec:Pruning}}

\vspace{-2mm}

Pruning the resultant DAG is common practice to suppress the redundant
edges \citep{Buhlmann_etall_14Cam,Zheng_etal_18DAGs,Wang_etal_2021Ordering,Bello_etal_22dagma,Duong_etal_24Alias},
and is also employed in our framework. This can be done by thresholding
the weight matrix for linear data \citep{Zheng_etal_18DAGs}, or employing
significance testing for nonlinear data using generalized additive
model regression \citep[CAM pruning, ][]{Peters_etal_2014Causal},
or conditional independence testing under the faithfulness assumption
\citep{Duong_etal_24Alias}. More details are provided in Appendix~\ref{subsec:Pruning-Techniques}.

\vspace{-3mm}

\section{Experiments\label{sec:Experiments}}

\begin{figure}[t]
\begin{centering}
\begin{tabular*}{1\columnwidth}{@{\extracolsep{\fill}}>{\centering}p{1\textwidth}}
\includegraphics[width=1\textwidth]{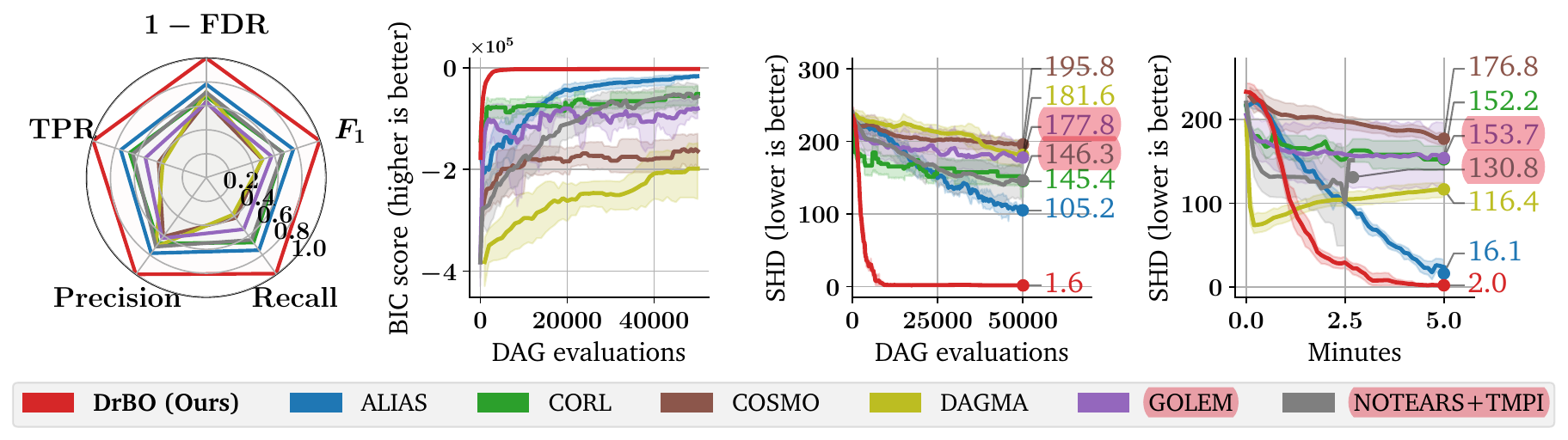}\tabularnewline
(a) \textbf{Dense graphs} with Linear-Gaussian data (DAGs with \textbf{30
nodes} and \textbf{\ensuremath{\approx}240 edges}). For fairness,
summary metrics (first column) are calculated at $50,\!000$ evaluations
for all methods.\tabularnewline
\includegraphics[width=1\textwidth]{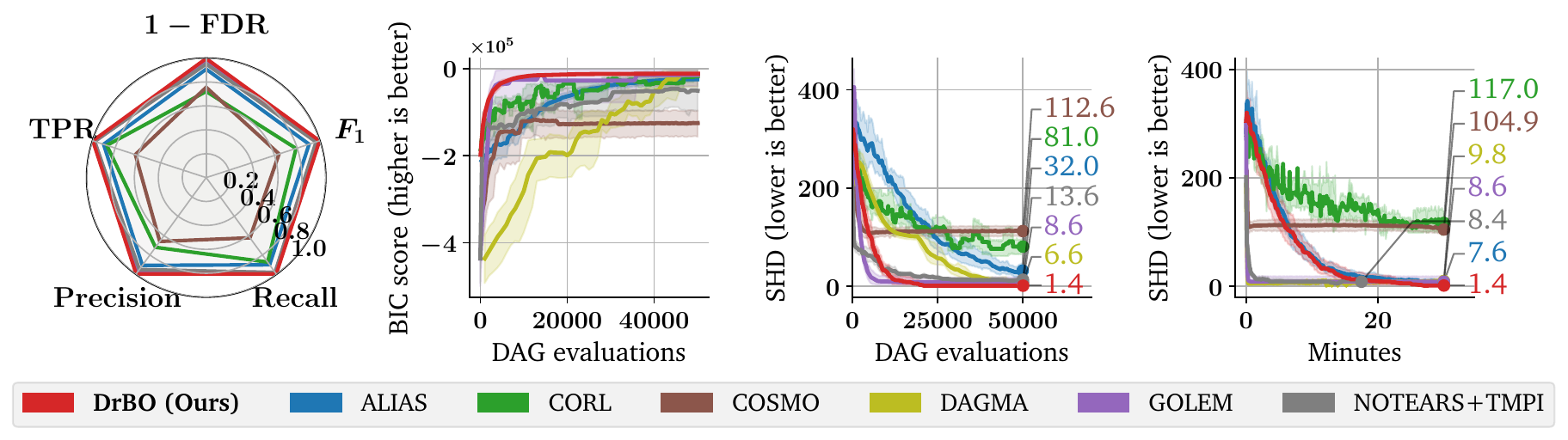}\tabularnewline
(b) \textbf{Large graphs} with Linear-Gaussian data (DAGs with \textbf{100
nodes} and \textbf{\ensuremath{\approx}200 edges}). For fairness,
summary metrics (first column) are calculated at $50,\!000$ evaluations
for all methods.\tabularnewline
\includegraphics[width=1\textwidth]{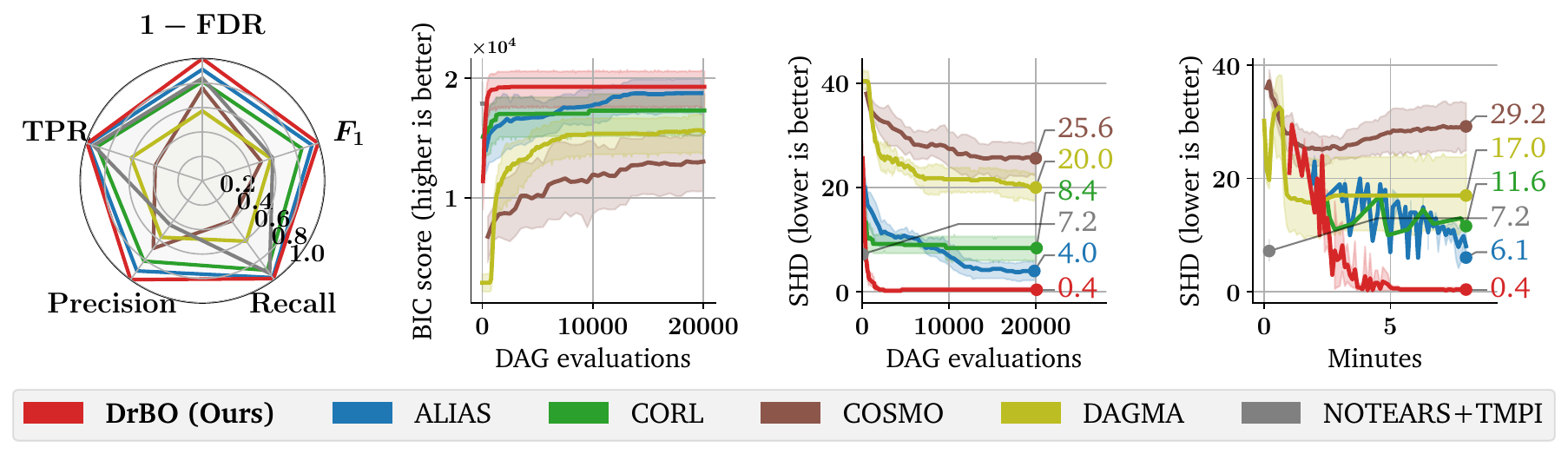}\tabularnewline
(c) \textbf{Nonlinear data} with Gaussian processes (DAGs with \textbf{10
nodes} and \textbf{\ensuremath{\approx}40 edges}). For fairness, summary
metrics (first column) are calculated at $20,\!000$ evaluations for
all methods.\tabularnewline
\end{tabular*}
\par\end{centering}
\caption{\textbf{DAG learning results on Synthetic data.} \textit{First column:}
overall performance in terms of True Positive Rate (TPR, higher is
better), Precision, Recall, and $F_{1}$ score (higher is better),
as well as False Discovery Rate (FDR, lower is better). \textit{Second
column:} we track the best Bayesian Information Criterion (BIC, higher
is better) so far at every optimization step. \textit{Third and Fourth
columns}: we monitor the Structural Hamming Distance (SHD, lower is
better) of the DAG whose best BIC so far at every optimization step.
Shaded areas in the line plots indicate 95\% confidence intervals
over 5 random datasets. NOTEARS+TMPI usually stops early before the
time limit.\label{fig:synthetic}}

\vspace{-4mm}
\end{figure}

\vspace{-1mm}

In this section, we verify our claim in the introduction: \textit{$\ours$
is both more accurate and sample-efficient than existing approaches}
\textit{in score-based observational DAG learning}. We show this by
comparing our $\ours$ method with a number of the most recent advances
in causal discovery that are based on sequential optimization, including
gradient-based methods DAGMA \citep{Bello_etal_22dagma}, COSMO \citep{Massidda_etal_2023Constraint},
GOLEM \citep{Ng_etal_2020Role}, NOTEARS \citep{Zheng_etal_18DAGs}
with TMPI constraint \citep{Zhang_etal_2022Truncated}, as well as
RL-based approaches CORL \citep{Wang_etal_2021Ordering} and ALIAS
\citep{Duong_etal_24Alias}. We note that CORL, ALIAS, and $\ours$
directly optimize the BIC score, COSMO, DAGMA, and NOTEARS optimize
the penalized least-square loss, while GOLEM optimizes a penalized
log-likelihood. For gradient-based methods, we consider a gradient
update equivalent to one DAG evaluation. Additional information, including
implementation details and metrics, are provided in Appendix~\ref{subsec:Experiment-Details}.

\vspace{-1mm}

\subsection{Results on Synthetic data}

\vspace{-1mm}

We consider the standard Erd\H{o}s-R\'{e}nyi (ER) graph model \citep{Erdos_Renyi_60Evolution}
to generate data, where graphs with $d$ nodes and $de$ edges on
average are referred to as $d$ER$e$ graphs (e.g, 10ER4).

\vspace{-1mm}

\subsubsection{Linear-Gaussian data\label{subsec:Linear-Gaussian-data}}

\vspace{-1mm}

After simulating a DAG $\mathcal{G}$, we sample edge weights with
$w_{ji}\sim\mathcal{U}\left(\left[-2,-0.5\right]\cup\left[0.5,2\right]\right)$
like prior studies \citep{Zheng_etal_18DAGs,Zhu_etal_2020Causal,Wang_etal_2021Ordering}.
Then, we generate a dataset of $n=1,\!000$ i.i.d. samples according
to a linear-Gaussian SCM $x_{i}:=\sum_{j\in\pa_{i}}w_{ji}x_{j}+\varepsilon_{i}$,
where $\varepsilon_{i}\sim\mathcal{N}\left(0,1\right)$. For fairness,
we prune the DAGs returned by $\ours$, along with ALIAS and CORL
as prescribed in their papers, by thresholding the weight matrix obtained
via linear regression at 0.3. This is not done for DAGMA and COSMO
because their implementations already incorporated the same pruning
scheme.

\textbf{Dense graphs.} We stress-test our method with complex structures
in Figure~\ref{fig:synthetic}(a). As depicted in the first column,
our method is the only approach that can achieve absolute overall
performance in all five metrics, surpassing the second- and third-best
methods ALIAS and CORL by large margins of more than 20\% in each
metric, while gradient-based methods DAGMA and COSMO struggle with
much worse performance. From the second and third columns, our $\ours$
approach can reach higher BIC scores, and thus lower SHDs, very sharply
with the number of evaluations, highlighting its sample-efficiency.
Lastly, the fourth column shows that our method's SHD improvement
over time is continuous and faster in terms of runtime than all other
baselines.

\textbf{Large-scale graphs.} Next, we demonstrate the scalability
of our method on high-dimensional graphs of 100 nodes, having up to
$10,\!000$ edges. Our results in Figure~\ref{fig:synthetic}(b)
shows that $\ours$ is still the leading method for high-dim data,
where it obtains absolute overall performance along with the lowest
SHD among all methods in limited runtime.

\vspace{-2mm}

\subsubsection{Nonlinear data\label{subsec:Nonlinear-data}}

\vspace{-2mm}

Following prior works \citep{Zhu_etal_2020Causal,Wang_etal_2021Ordering,Yang_etal_2023Reinforcement,Duong_etal_24Alias},
we demonstrate our method on nonlinear datasets generated using Gaussian
processes from \citet{Lachapelle_etal_2020Gradient}. Specifically,
we employ the datasets generated according to an ANM $x_{i}:=f_{i}\left(\mathbf{x}_{\pa_{i}}\right)+\varepsilon_{i}$
where $f_{i}$ is drawn from a Gaussian process prior with a unit
bandwidth RBF kernel, and $\varepsilon_{i}\sim\mathcal{N}\left(0,\sigma_{i}^{2}\right)$
with non-equal noise variances $\sigma_{i}^{2}$ sampled uniformly
in $\left[0.4,0.8\right]$. ALIAS, CORL, and $\ours$ optimize $S_{\mathrm{BIC-NV}}$,
while COSMO, DAGMA, and NOTEARS+TMPI optimize the least-square objective.
The empirical results reported in Figure~\ref{fig:synthetic}(c)
confirms that our $\ours$ method also excels on complex nonlinear
data in both accuracy and computational cost. This is evidenced by
the leading overall performance (first column), along with a vanishing
SHD of only $0.4$ at the end of the learning curve, surpassing other
methods by a visible gap both in SHD and convergence speed.

\begin{wrapfigure}{o}{0.4\columnwidth}%
\vspace{-6mm}
\includegraphics[width=0.4\textwidth]{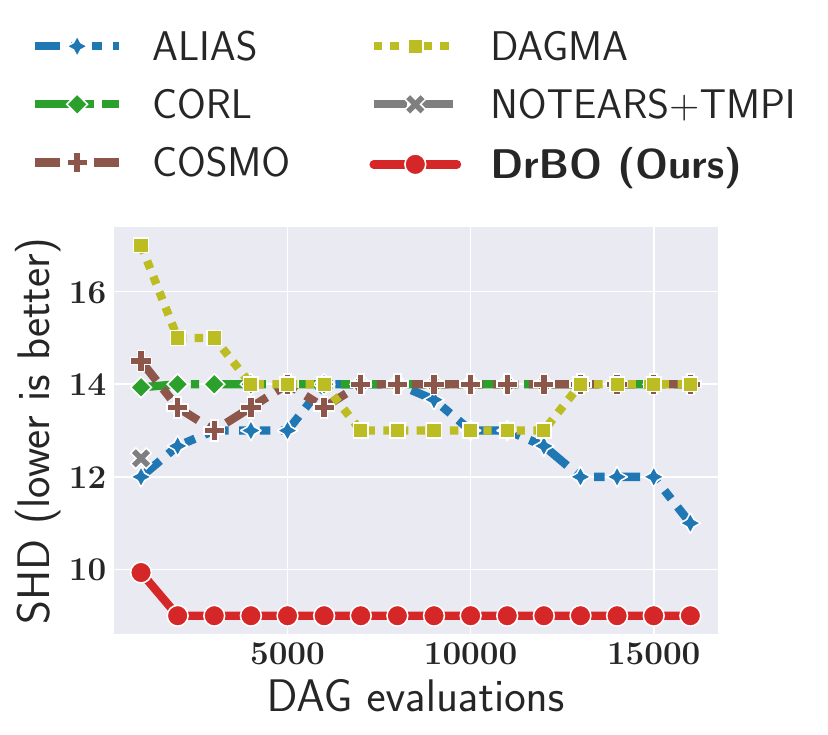}

\vspace{-3mm}

\caption{\textbf{Causal Discovery Performance on the Benchmark Sachs Dataset}
\textbf{\citep{Sachs_etall_05Causal}}. NOTEARS+TMPI stops early before
the max no. of evaluations is reached.\label{fig:sachs}}

\vspace{-5mm}
\end{wrapfigure}%

\vspace{-2mm}

\subsection{Results on Real Data and Structures}

\vspace{-5mm}

\begin{table}[h]
\caption{\textbf{Causal Discovery Performance on Real-world Structures \citep{Scutari_10Learning}}.
The performance is measured in Structural Hamming Distance (SHD, lower
is better). The numbers are $\mathrm{mean}\pm\mathrm{std}$ over 5
independent datasets with $1,\!000$ observational samples. For fairness,
all methods are limited to $20,\!000$ evaluations.\label{tab:bnlearn-results}}

\resizebox{\textwidth}{!}{

\begin{tabular}{cccccc}
\toprule 
\textbf{Dataset} & \textbf{Alarm} & \textbf{Asia} & \textbf{Cancer} & \textbf{Child} & \textbf{Earthquake}\tabularnewline
\textbf{Method} & \textbf{(37 nodes, 46 edges)} & \textbf{(8 nodes, 8 edges)} & \textbf{(5 nodes, 4 edges)} & \textbf{(20 nodes, 25 edges)} & \textbf{(5 nodes, 4 edges)}\tabularnewline
\midrule
ALIAS \citep{Duong_etal_24Alias} & $26.8\pm7.8$ & $0.2\pm0.5$ & $0.0\pm0.0$ & $\phantom{0}2.8\pm1.6$ & $\mathbf{0.0\pm0.0}$\tabularnewline
CORL \citep{Wang_etal_2021Ordering} & $19.8\pm8.6$ & $\mathbf{0.0\pm0.0}$ & $\mathbf{0.0\pm0.0}$ & $\phantom{0}1.6\pm2.3$ & $\mathbf{0.0\pm0.0}$\tabularnewline
COSMO \citep{Massidda_etal_2023Constraint} & $26.8\pm4.0$ & $4.0\pm1.6$ & $2.2\pm0.8$ & $11.6\pm2.3$ & $2.2\pm0.8$\tabularnewline
DAGMA \citep{Bello_etal_22dagma} & $25.0\pm5.3$ & $2.6\pm1.3$ & $1.0\pm1.2$ & $\phantom{0}8.0\pm5.2$ & $1.0\pm1.2$\tabularnewline
GOLEM \citep{Ng_etal_2020Role} & $\phantom{0}4.8\pm6.2$ & $\mathbf{0.0\pm0.0}$ & $\mathbf{0.0\pm0.0}$ & $\mathbf{0.0\pm0.0}$ & $\mathbf{0.0\pm0.0}$\tabularnewline
NOTEARS+TMPI \citep{Zheng_etal_18DAGs,Zhang_etal_2022Truncated} & $\phantom{0}7.0\pm7.9$ & $0.4\pm0.9$ & $\mathbf{0.0\pm0.0}$ & $\mathbf{0.0\pm0.0}$ & $\mathbf{0.0\pm0.0}$\tabularnewline
\midrule 
$\ours$ (Ours) & $\mathbf{\phantom{0}1.0\pm2.2}$ & $\mathbf{0.0\pm0.0}$ & $\mathbf{0.0\pm0.0}$ & $\phantom{0}\mathbf{0.0\pm0.0}$ & $\mathbf{0.0\pm0.0}$\tabularnewline
\bottomrule
\end{tabular}

}
\end{table}

\vspace{-3mm}

\textbf{Benchmark data.} We verify the performance of our method on
real data using the popular benchmark flow cytometry dataset \citep{Sachs_etall_05Causal},
concerning a protein signaling network based on expression levels
of proteins and phospholipids. We employ the observational portion
of the dataset containing 853 observations and a known causal network
with 11 nodes and 17 edges. We apply the similar settings as in Sec.~\ref{subsec:Nonlinear-data}
for all methods. The evaluations shown in Figure~\ref{fig:sachs}
verify the effectiveness of our method on real data, where it effortlessly
achieves a lowest SHD of 9 with fewer evaluations compared with the
competitors.

\textbf{Real-world structures.} To further illustrate the capabilities
of our approach on real-world scenarios, we conduct experiments on
real structures provided by the BnLearn repository \citep{Scutari_10Learning}.
Each dataset contains $1,\!000$ observational samples and a ground
truth causal network belonging to real-world applications with varying
size. Additional details regarding these datasets are given in Appendix~\ref{subsec:BnLearn-Structures}.
The results are presented in Table~\ref{tab:bnlearn-results}, highlighting
that our method is the only approach that can consistently achieve
zero SHD on four out of five real-world structures. On the Alarm dataset,
which appears to be most challenging, our $\ours$ method still leads
with much lower SHD compared with all other baselines.

\vspace{-3mm}

\subsection{Supplementary Results}

\vspace{-2mm}

\textbf{Extended Causal Discovery Settings.} We investigate the performance
of $\ours$ in extended scenarios in Appendix~\ref{subsec:Additional-Settings}
as follows.\textbf{ Varying sample sizes:} we show in Figure~\ref{fig:sample-size}
that our method can achieve low SHDs even with limited data.\textbf{
Different graph models:} in Table~\ref{tab:different-graphs}, $\ours$
achieves low SHDs and surpasses competitors for both ER and SF graphs,
even on the dense graphs.\textbf{ Different noise distributions:}
Table~\ref{tab:different-noises} shows that $\ours$ also outperforms
the baselines under five different noise types.\textbf{ BGe score:}
Figure~\ref{fig:bge} confirms that our method can work with a different
score and match the score of ground truth graphs with low structural
errors.\textbf{ Discrete Data:} in Figure~\ref{fig:Discrete-data},
we show that $\ours$ also obtains the highest scores and lowest SHDs
compared with the baselines for non-continuous data. \textbf{Standardized
Data}: we show in Figure~\ref{fig:standardized-result} that our
method is also robust against data standardization for both linear
and nonlinear data. \textbf{Large-scale Nonlinear Data:} Figure~\ref{fig:nonlinear-large}
demonstrates $\ours$'s competitive performance and efficiency for
nonlinear data on 50- and 100-node graphs.

\vspace{-1mm}

\textbf{Additional baselines.} In Appendix~\ref{subsec:additional-baselines},
we also examine other baselines that are not based on sequential optimization,
showing that $\ours$ also significantly outperforms these methods
on linear, nonlinear, as well as real data.

\textbf{Runtime.} In Appendix~\ref{subsec:Time-Comparisons}, we
also detail the numerical runtime among all methods.

\textit{}

\vspace{-8mm}

\subsection{Ablations}

\vspace{-1mm}

\begin{figure}[t]
\resizebox{1\textwidth}{!}{

\begin{tabular}{>{\centering}p{0.5\columnwidth}>{\centering}p{0.5\linewidth}}
\includegraphics[width=0.51\columnwidth]{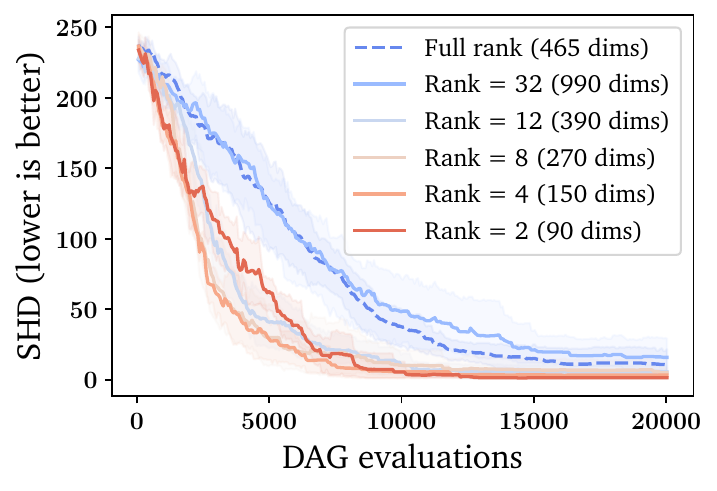} & \includegraphics[width=0.51\columnwidth]{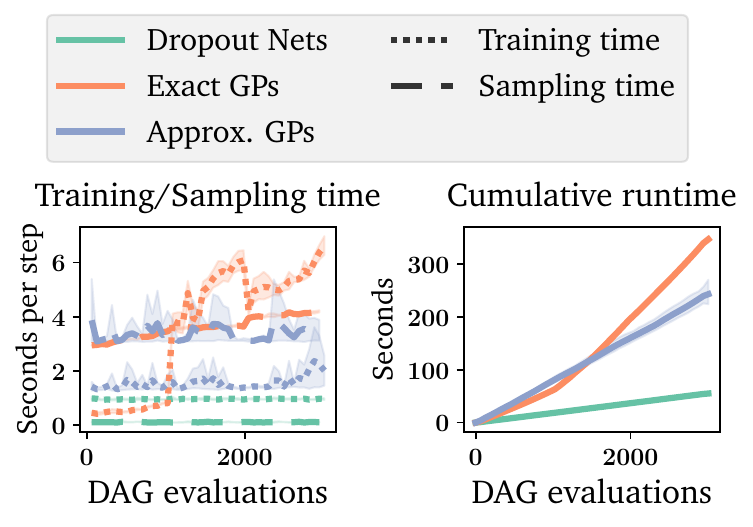}\tabularnewline
\textbf{(a) Effect of Low-Rank Representation.} We compare this for
different ranks $k$ with the Full-rank representation \citep{Duong_etal_24Alias}.
See our discussion in Appendix~\ref{subsec:Why-low-rank-DAG}. & \textbf{(b) Effect of Dropout Networks.} We compare these with Exact
GPs \citep{Rasmussen_03Gaussian} and Approximate GPs \citep[AGP,][]{Hensman_etal_15Scalable}.
Details in Appendix~\ref{subsec:Effect-of-Dropout}.\tabularnewline
\includegraphics[width=0.51\columnwidth]{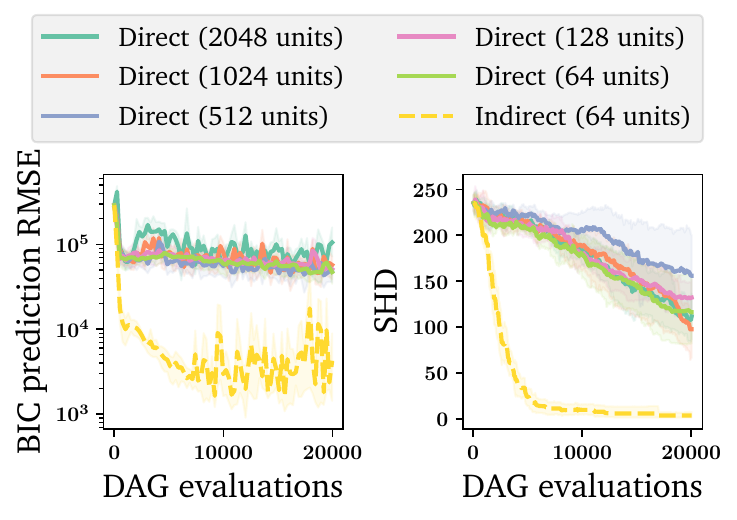} & \includegraphics[width=0.51\columnwidth]{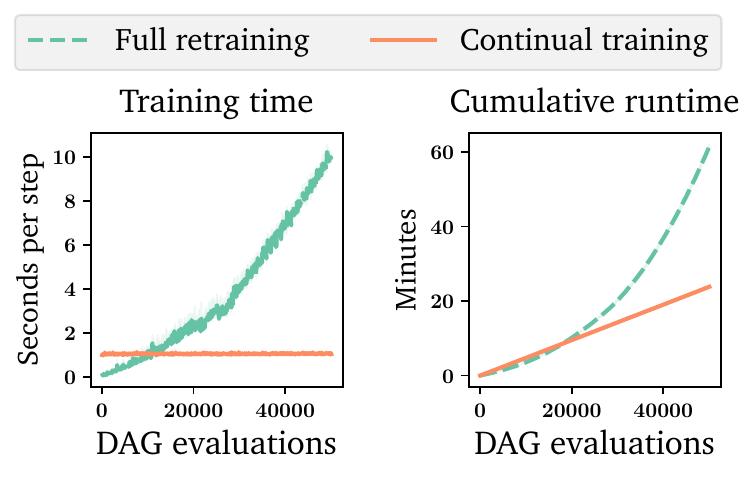}\tabularnewline
\textbf{(c) Effect of Indirect DAG Modeling.} We compare this with
learning a Direct map from a DAG to its score, with varying no. hidden
units. & \textbf{(d) Effect of Continual Training.} We compare this approach
with Full retraining using all experienced data at every BO iteration.
Details in Appendix~\ref{subsec:Effect-of-Continual}.\tabularnewline
\end{tabular}

}

\vspace{-2mm}

\caption{\textbf{Ablating our design choices.} All configurations are evaluated
on 5 linear-Gaussian datasets of $1,\!000$ samples on 30ER8 graphs.
Shaded areas indicate 95\% confidence intervals.\label{fig:Ablating}}

\vspace{-3mm}
\end{figure}

In Figure~\ref{fig:Ablating}, through ablation studies, we justify
the design choices outlined in the Introduction, showing that every
component contributes considerately to the accuracy and/or scalability
of our method. The remaining hyperparameters of our algorithm are
also studied in Appendix~\ref{subsec:Ablation-Experiments}.

\vspace{-3mm}

\section{Concluding Remarks\label{sec:Conclusion}}

\vspace{-2mm}

This study presents $\ours$, a novel BO method to search for high-scoring
DAGs. We have shown that, by meticulously choosing promising DAGs
to evaluate, we can find the optimal one more efficiently and cost-effectively.
Our comprehensive experiments demonstrate that $\ours$ performs well
even in many intricate settings like dense graphs, high-dimensional,
and nonlinear data.

Regarding limitations, in our method, the surrogate model architecture
is manually chosen and remains fixed through the course of optimization,
and thus is prone underfitting at the end. To mitigate this, incremental
neural architecture search techniques \citep{Liu_etal_18Progressive,Geifman_ElYaniv_19Deep}
can be employed to facilitate autonomous architecture selection and
scaling. In addition, our continual learning component is simple and
can benefit from more advanced techniques \citep{Wang_etal_24comprehensive}
to further improve the performance of our surrogate models.

For future developments, it would be an interesting direction to combine
our approach with active causal discovery methods to recover the causal
structure even more efficiently. Moreover, our method can be extended
to solve causal discovery problems with hidden confounders, where
the outputs are no longer DAGs.
\vfill{}

\pagebreak{}

\bibliographystyle{iclr2025_conference}
\bibliography{ref}
\newpage{}

\appendix

\section{Proofs}

\subsection{Proof of Lemma~\ref{lem:vec2dag-lr}\label{subsec:proof-vec2dag-lr}}
\begin{proof}
\noindent The acyclicity of Eq.~(\ref{eq:vec2dag-lr}) is ensured
by the first term: $H\left(\mathrm{grad}\left(\mathbf{p}\right)\right)$.
This is a binary matrix representing the adjacency matrix of a directed
graph. In this graph, the presence of an edge $i\rightarrow j$ is
equivalent to $p_{i}<p_{j}$. As such, a cycle $i_{1}\rightarrow\ldots\rightarrow i_{1}$,
if exists, would lead to $p_{i}<p_{i}$, which is contradictory, meaning
this graph must be a DAG. Multiplying this adjacency matrix with the
second term $H\left(\mathbf{R}\cdot\mathbf{R}^{\top}\right)$, which
is also a binary matrix, has the effect of removing already existing
edges in this graph, and thus cannot introduce any cycle. This concludes
our proof.
\end{proof}

\subsection{Proof of Lemma~\ref{lem:scale-invariance}\label{Proof-of-scale-invariance}}
\begin{proof}
For any $\alpha>0$, we have $H\left(p_{i}-p_{j}\right)=H\left(\alpha\left(p_{i}-p_{j}\right)\right)=H\left(\alpha p_{i}-\alpha p_{j}\right)$
and $H\left(\mathbf{R}\cdot\mathbf{R}^{\top}\right)=H\left(\alpha^{2}\mathbf{R}\cdot\mathbf{R}^{\top}\right)=H\left(\left(\alpha\mathbf{R}\right)\cdot\left(\alpha\mathbf{R}\right)^{\top}\right)$.
Thus, $H\left(\mathrm{grad}\left(\mathbf{p}\right)\right)\odot H\left(\mathbf{R}\cdot\mathbf{R}^{\top}\right)=H\left(\mathrm{grad}\left(\alpha\mathbf{p}\right)\right)\odot H\left(\left(\alpha\mathbf{R}\right)\cdot\left(\alpha\mathbf{R}\right)^{\top}\right)$,
concluding our proof.
\end{proof}

\section{Deriving BIC Scores\label{subsec:Derivation-of-BIC}}

Recall that for a causal model with parameters $\theta:=\left\{ \left\{ f_{i}\right\} _{i=1}^{d},P\left(\bm{\varepsilon}\right)\right\} $,
the general BIC is given by 
\begin{equation}
S_{\text{BIC}}\left(\mathcal{D},\mathcal{G}\right):=2\ln p\left(\mathcal{D}\mid\hat{\theta},\mathcal{G}\right)-\left|\mathcal{G}\right|\ln n,\label{eq:bic}
\end{equation}
where $\hat{\theta}:=\arg\max_{\theta}p\left(\mathcal{D}\mid\theta,\mathcal{G}\right)$
is the maximum-likelihood estimator of the causal model parameters,
$n$ is the sample size of $\mathcal{D}$, and $\left|\mathcal{G}\right|$
denotes the number of edges in $\mathcal{G}$. The first term in Eq.~(\ref{eq:bic})
is a log-likelihood objective similar to GOLEM \citep{Ng_etal_2020Role},
GraN-DAG \citep{Lachapelle_etal_2020Gradient}, etc., while the second
term penalizes extra edges.

The flexibility of Eq.~(\ref{eq:bic}) allows for the adoption of
BIC in various causal models by simply specifying the likelihood model.
In the following, we derive the BIC scores for additive noise models
with non-equal and equal variances, as well as logistic models.

\subsection{BIC for ANM with Non-qual Variances}

Let us consider an ANM defined by
\begin{equation}
x_{i}:=f_{i}\left(\mathbf{x}_{\pa_{i}^{\mathcal{G}}}\right)+\varepsilon_{i},\ \forall i=1,\ldots,d,\label{eq:anm}
\end{equation}

where the noise is assumed to be Gaussian with fixed variance: $\varepsilon_{i}\sim\mathcal{N}\left(0,\sigma_{i}^{2}\right)\ \forall i=1,\ldots d$.
The likelihood of a dataset $\mathcal{D}=\left\{ \mathbf{x}^{\left(j\right)}\right\} _{j=1}^{n}$
under this model is given by
\begin{equation}
\mathcal{L}=\ln p\left(\mathcal{D}\mid\mathcal{G},\left\{ f_{i}\right\} _{i=1}^{d},\left\{ \sigma_{i}\right\} _{i=1}^{d}\right)=-\frac{1}{2}\sum_{i=1}^{d}\frac{\sum_{j=1}^{n}\left(x_{i}^{\left(j\right)}-f_{i}\left(\mathbf{x}_{\pa_{i}^{\mathcal{G}}}^{\left(j\right)}\right)\right)^{2}}{\sigma_{i}^{2}}-\frac{n}{2}\sum_{i=1}^{d}\ln\sigma_{i}^{2}+\text{constant}.\label{eq:log-likelihood}
\end{equation}

Taking its derivative and setting it to zero, we can solve for the
parameters as follows: 
\begin{equation}
\hat{\sigma}_{i}^{2}=\underbrace{\frac{1}{n}\sum_{j=1}^{n}\left(x_{i}^{\left(j\right)}-\hat{f_{i}}\left(\mathbf{x}_{\pa_{i}^{\mathcal{G}}}^{\left(j\right)}\right)\right)^{2}}_{\text{MSE}_{i}},
\end{equation}

where $\left\{ \hat{f_{i}}\right\} _{i=1}^{d}$ are least-square estimators
obtained by minimizing $\left(x_{i}^{\left(j\right)}-f_{i}\left(\mathbf{x}_{\pa_{i}^{\mathcal{G}}}^{\left(j\right)}\right)\right)^{2}$
over a restricted hypothesis class $f\sim\mathcal{F}$. Following
past studies \citep{Zhu_etal_2020Causal,Wang_etal_2021Ordering,Yang_etal_2023Reinforcement,Yang_etal_2023Causal,Duong_etal_24Alias},
we use linear regression for linear data and Gaussian process regression
for nonlinear data. However, any other regression method can be used
as desired.

Subsequently, the first term in Eq.~(\ref{eq:log-likelihood}) cancels
out and the maximum log-likelihood is reduced to
\begin{equation}
\mathcal{\hat{L}}=-\frac{n}{2}\sum_{i=1}^{d}\ln\text{MSE}_{i}+\text{constant}.
\end{equation}

Finally, the BIC score for ANMs with non-equal variances is obtained
by:

\begin{equation}
S_{\text{BIC-NV}}\left(\mathcal{D},\mathcal{G}\right)=-n\sum_{i=1}^{d}\ln\text{MSE}_{i}-\left|\mathcal{G}\right|\ln n.
\end{equation}

\subsection{BIC for ANM with Equal Variances}

By setting $\sigma_{i}=\sigma\ \forall i=1,\ldots,d$, we repeat the
previous steps and obtain the following solution for the maximum likelihood
estimators: 
\begin{equation}
\hat{\sigma}^{2}=\frac{1}{d}\sum_{i=1}^{d}\underbrace{\frac{1}{n}\sum_{j=1}^{n}\left(x_{i}^{\left(j\right)}-\hat{f}_{i}\left(\mathbf{x}_{\pa_{i}}^{\left(j\right)}\right)\right)^{2}}_{\text{MSE}_{i}}.
\end{equation}

The maximum likelihood now is given by
\begin{equation}
\hat{\mathcal{L}}=-\frac{nd}{2}\ln\frac{\sum_{i=1}^{d}\text{MSE}_{i}}{d}+\text{constant}.
\end{equation}

Finally, we obtain the the BIC score for ANMs with equal variances
as follows:

\begin{equation}
S_{\text{BIC-EV}}\left(\mathcal{D},\mathcal{G}\right)=-nd\ln\frac{\sum_{i=1}^{d}\text{MSE}_{i}}{d}-\left|\mathcal{G}\right|\ln n.
\end{equation}

\subsection{BIC for Binary Data with Logistic Regression}

From the formulation in Eq.~(\ref{eq:bic}), we can also adapt it
to non-continuous data. For example, let us consider the logistic
causal model governed by
\begin{equation}
x_{i}\sim\text{Bernoulli}\left(f_{i}\left(\mathbf{x}_{\pa_{i}^{\mathcal{G}}}\right)\right).\label{eq:logistic}
\end{equation}

where $f_{i}$ models the conditional probability of $x_{i}$ given
$\mathbf{x}_{\pa_{i}^{\mathcal{G}}}$.

The log-likelihood of data under this model is determined by
\[
\mathcal{L}=\ln p\left(\mathcal{D}\mid\mathcal{G},\left\{ f_{i}\right\} _{i=1}^{d}\right)=\sum_{i=1}^{d}\sum_{j=1}^{n}\left(x_{i}^{\left(j\right)}\ln f_{i}\left(\mathbf{x}_{\pa_{i}^{\mathcal{G}}}^{\left(j\right)}\right)+\left(1-x_{i}^{\left(j\right)}\right)\ln\left(1-f_{i}\left(\mathbf{x}_{\pa_{i}^{\mathcal{G}}}^{\left(j\right)}\right)\right)\right).
\]

Maximizing this objective and plugging it into Eq.~(\ref{eq:bic})
gives us with the BIC for logistic model:

\begin{equation}
S_{\text{BIC-Logistic}}\left(\mathcal{D},\mathcal{G}\right)=2\hat{\mathcal{L}}-\left|\mathcal{G}\right|\ln n.\label{eq:bic-logistic}
\end{equation}

The first term is equivalent to the logistic loss used in, e.g., DAGMA
\citep{Bello_etal_22dagma}, while the second term punishes redundant
edges as always.

\section{Additional Discussions and Details}

\subsection{Preliminary Candidate Generation\label{sec:Latin-Hypercube-Design}}

In Sec.~\ref{subsec:Acquisition-Function-Optimizatio}, we generate
a set of $C$ preliminary candidates in a hyperrectangle centered
at the best solution so far $\mathbf{z}^{\ast}$. This hyperrectangle
can be seen as a single ``trust region'' in BO \citep{Eriksson_etal_19Scalable,Daulton_etal_22Multi}.
More specifically, our trust region is the intersection of the search
space $\left[-1,1\right]^{d\left(1+k\right)}$ and the hypercube of
length $L$ centered at $\mathbf{z}^{\ast}$. Following \citet{Eriksson_etal_19Scalable},
we adaptively update $L$ according to the learning progress. This
is done by maintaining a success (and failure) counter that keeps
track the number of consecutive BO iterations that improves (or fails
to improve, resp.) the DAG score. After $n_{\text{succ}}$ consecutive
successes, we enlarge $L$ by two times in order to shift the focus
to other regions, and after $n_{\text{fail}}$ consecutive failures,
we shrink it by two times to zoom more into the current region. In
all experiments, we use the fixed values of $n_{\text{succ}}=3$,
$n_{\text{fail}}=5$, and $L$ is initialized with the value of $1$
and is clipped to be within $\left[0.01,2\right]$ after each update.

To produce the preliminary candidates on which we generate Thompson
samples, for the very first suggestions, following \citet{Eriksson_etal_19Scalable},
we employ the Latin hypercube design \citep[LHD,][]{McKay_etal_00A},
which is a space-filling method used to generate near-random samples
from a multidimensional space, which ensures that each dimension of
the hypercube is evenly covered, while random sampling can lead to
an unevenly covered space. For subsequent suggestions, we follow the
established procedure in \citep{Eriksson_etal_19Scalable} to first
generate a scrambled Sobol sequence \citep{Owen_98Scrambling} within
the current trust region, then we use the perturbation value in the
Sobol sequence with probability $\min\left\{ 1,\frac{20}{d\left(1+k\right)}\right\} $
for each given candidate and dimension, and the value of the center
$\mathbf{z}^{\ast}$ otherwise. As noted in \citep{Regis_Shoemaker_13Combining,Eriksson_etal_19Scalable},
perturbing only a few dimensions can lead to a significant performance
improvement for high-dim scenarios.

\subsection{Scalable Surrogate Modeling\label{sec:Discussion-on-scalable-surrogate}}

The literature of BO is vast and here we only discuss a few promising
alternative approaches to scale up surrogate models in BO, and justify
of our dropout neural network choice.

\textbf{Bayesian neural networks} \textbf{(BNN)} are a also natural
replacement thanks to the flexibility of neural networks combined
with the inherent ability to model uncertainty of the Bayesian ideology
\citep{Springenberg_etal_16Bayesian}. However, to stay as close as
possible to a truly Bayesian treatment, i.e., characterizing the exact
posterior distributions, they require stochastic gradient Markov Chain
Monte Carlo (MCMC) to sample from the posterior, which necessitates
several sampling steps to reach the desirable posterior \citep{Springenberg_etal_16Bayesian}.
Meanwhile, in our method, we sacrifice the accuracy of the posterior
inference, so that we can sample from the posterior faster with a
single forward pass through the dropout neural networks. That being
said, our DAG learning accuracy is still strong, which justifies this
sacrifice.

\textbf{Sparse GPs (or Approximate GPs)} have also been studied to
enhance scalability of GPs \citep{Snelson_05Sparse,Hensman_etal_15Scalable,Titsias_09Variational}.
However, they do not scale well with dimensionality \citep{Wang_18Batched}.

\textbf{Random forest (RF)} has also been considered as an alternative
surrogate model in BO \citep{Hutter_etal_11Sequential}. An RF is
composed of multiple decision trees, each of which is constructed
from a portion of the training set and a few random dimensions. Therefore,
uncertainty modeling in RF is achieved via the variation of the individual
trees' predictions. While RF is known to be efficient, especially
in tabular data, we find it more straightforward to train neural networks
continually \citep{Wang_etal_24comprehensive} compared with tree-based
models \citep{Utgoff_89Incremental,Utgoff_97Decision}. This is crucial
for our BO framework to scale with the number of iterations as aforementioned.

\textbf{Ensembling methods} \citep{Wang_18Batched,Guo_etal_18Heterogeneous}
sidestep the scalability challenge of BO with the use of an ensemble
of models trained on different partitions of the samples and dimensions.
A similar idea is also proposed in \citep{Eriksson_etal_19Scalable},
where local GPs are trained on multiple local trust regions of the
search space. However, these methods require training multiple models,
and managing them is less straightforward than maintaining a single
neural network like our method, which also scales very well with a
strong causal discovery performance.

\subsection{Pruning Techniques\label{subsec:Pruning-Techniques}}

Pruning is typically employed in causal discovery to reduce false
positive estimates, and there are several approaches depending on
the causal model as follows.

\textbf{Linear data.} For linear data, given the resultant DAG $\mathcal{G}$
that needs to be pruned, linear regression is used to find the linear
coefficients $\left\{ \hat{w}_{ij}\right\} _{\left(i\rightarrow j\right)\in\mathcal{G}}$
associating with the edges in $\mathcal{G}$. Then, only the weights
satisfying a certain absolute strength $\alpha$ is kept, while the
remaining are removed. In other words, the edge $\left(i\rightarrow j\right)\in\mathcal{G}$
is kept iff $\left|\hat{w}_{ij}\right|>\alpha$. Typically, the weights
in the true generative process for linear models are sampled from
$\mathcal{U}\left(\left[-2,-0.5\right]\cup\left[0.5,2\right]\right)$,
and the common value for $\alpha$ is $0.3$ \citep{Zheng_etal_18DAGs,Zhu_etal_2020Causal,Wang_etal_2021Ordering,Bello_etal_22dagma,Massidda_etal_2023Constraint,Duong_etal_24Alias}.

\textbf{Nonlinear additive model.} For nonlinear data under additive
models, a common approach is based on feature selection using generalized
additive model (GAM) regression \citep{Buhlmann_etall_14Cam}, also
known as CAM pruning. Particularly, each node $i$ is regressed on
its parents $\mathbf{x}_{\pa_{i}^{\mathcal{G}}}$ using GAM, then
the significance test of covariates is conducted, and a parent is
kept if its $p$-value is lower than the significance level of $0.001$.
This is also done in \citep{Zhu_etal_2020Causal,Wang_etal_2021Ordering,Massidda_etal_2023Constraint,Rolland_etal_22Score,Sanchez_etal_22Diffusion,Duong_etal_24Alias}.

\textbf{Non-additive models.} For general non-additive models, conditional
independence (CI) testing can also be employed to prune edges. To
be more specific, the Faithfulness assumption \citep{Peters_etal_17Elements}
implies that any conditional independence observed in data reflects
the corresponding $d$-separation in the causal graph, so if $x_{i}\indep x_{j}\mid\mathbf{x}_{\pa_{i}\setminus\left\{ j\right\} }$
for some $j\in\pa_{i}$, then $j$ is an extra edge and needs to be
removed. This follows the same idea of constraint-based causal discovery
\citep{Spirtes_etal_00Causation,Colombo_etal_12Learning} and feature
selection via Markov Blankets \citep{Kollar_Sahami_96optimal,Xing_etal_01Feature}.
As observed in \citep{Duong_etal_24Alias}, CI-based pruning leads
to better performance on the Sachs dataset compared with CAM pruning,
and hence we also employ it for all methods on the Sachs dataset.
The specific CI test is the popular kernel-based method KCIT \citep{Zhang_etal_11Kernel}
with a significance level of $0.001$.

\section{Experiment Details\label{subsec:Experiment-Details}}

\subsection{DAG Learning Metrics\label{subsec:Metrics}}

We evaluate the performance of each method using the following standard
measures, each of which calculates the disparity between an estimated
DAG and the ground truth DAG:
\begin{itemize}
\item \textbf{Structural Hamming Distance (SHD, lower is better):} this
is the most common metric in causal discovery, which counts the minimum
number of edge additions, removals, and reversals to turn the estimated
DAG into the true graph.
\item \textbf{BIC score (higher is better):} we also monitor the BIC score
of the estimated DAG for all methods in the main text (despite the
fact that some baselines do not optimize this score). For linear data,
we use the BIC with equal variances and linear regression, while for
nonlinear data, BIC with non-equal variance with Gaussian process
regression is used, and BIC with logistic regression is employed for
binary data.
\item \textbf{True Positive Rate (TPR, higher is better):} this measures
the ratio of correctly recovered edges over the true edges in the
ground truth DAG.
\item \textbf{False Discovery Rate (FDR, lower is better):} this measures
the proportion of incorrectly estimated edges over all estimated edges.
\item \textbf{Precision, Recall, and $F_{1}$}: this measures the binary
classification performance by treating the binary adjacency matrix
as a set of individual binary classification tasks.
\end{itemize}

\subsection{Synthetic Causal Graphs}

Our synthetic causal graphs involve one of the two well-known graph
models:
\begin{itemize}
\item \textbf{Erd\H{o}s-R\'{e}nyi} \citep[ER,][]{Erdos_Renyi_60Evolution}:
the edges in this type of graph are added independently with fixed
probability. To generate a DAG of $d$ nodes with an expected in-degree
of $e$, we first generate an undirected graph where edges are added
with probability $\frac{4de}{d\left(d-1\right)}$, then orient the
edges using a random permutation over the list of nodes.
\item \textbf{Scale-Free} \citep[SF,][]{Barabasi_Albert_99Emergence}: these
are graphs where the degree distribution follows a power law, where
a few nodes have many connections while others have only a few connections.
To generate SF graphs with $\approx de$ edges, we start with an empty
graph then repeatedly grow it by attaching new nodes, each with $k$
edges, that are preferentially attached to existing nodes.
\end{itemize}
\begin{rem}
It is noteworthy that the majority of methods perform well only for
graphs with up to $e=4$ \citep{Zheng_etal_18DAGs,Zhu_etal_2020Causal,Ng_etal_2020Role,Yu_etal_2021Dags,Wang_etal_2021Ordering,Bello_etal_22dagma,Yang_etal_2023Causal}.
Indeed, it was noted in \citep{Bello_etal_22dagma} that ER4 and SF4
are the hardest settings. However, in this study, we have shown that
our $\ours$ method is still very accurate on much denser ER8 and
SF8 graphs.
\end{rem}

\subsection{BnLearn Datasets\label{subsec:BnLearn-Structures}}

The BnLearn repository\footnote{Data is publicly downloadable at \url{https://www.bnlearn.com/bnrepository}}
\citep{Scutari_10Learning} contains a set of Bayesian networks of
varying sizes and complexities from different real-world domains.
The chosen networks in our study include Alarm \citep{Beinlich_etal_89ALARM},
Asia \citep{Lauritzen_88Local}, Cancer \citep{Korb_Nicholson_10Bayesian},
Child \citep{Spiegelhalter_92Learning}, and Earthquake \citep{Korb_Nicholson_10Bayesian}.
However, these datasets only describe the conditional probability
distributions of discrete variables, while the baselines considered
in our method are mostly implemented for continuous data. Since the
purpose of this experiment is to show that \textit{our method can
correctly recover real structures}, we use the networks from the BnLearn
to generate continuous data. Specifically, we employ linear-Gaussian
SCMs as in Sec.~\ref{subsec:Linear-Gaussian-data} to generate synthetic
data adhering to real causal networks, and each dataset contains only
$1,\!000$ observational samples.

\subsection{Implementations and Platform\label{subsec:Implementation-Details}}

\textbf{Implementations.} We employ the following implementations
for the considered baselines as follows:
\begin{itemize}
\item DAGMA \citep{Bello_etal_22dagma}: we use the official implementation
provided by the authors at \url{https://github.com/kevinsbello/dagma}.
\item COSMO \citep{Massidda_etal_2023Constraint}: we use the original implementation
attached as supplementary material at \url{https://openreview.net/forum?id=KWO8LSUC5W}.
\item CORL \citep{Wang_etal_2021Ordering}: we employ the official implementation
provided in the gCastle library \citep{Zhang_etal_2021Gcastle} at
\url{https://github.com/huawei-noah/trustworthyAI}.
\item ALIAS \citep{Duong_etal_24Alias}: we reimplement their method by
following the exact instructions and libraries described.
\item GOLEM \citep{Ng_etal_2020Role}: we adopt the implementation provided
by gCastle \citep{Zhang_etal_2021Gcastle} at \url{https://github.com/huawei-noah/trustworthyAI}.
\item NOTEARS+TMPI \citep{Zheng_etal_18DAGs,Zhang_etal_2022Truncated}:
we replace the DAG constraint in NOTEARS' implementation at \url{https://github.com/xunzheng/notears}
with the TMPI constraint at \url{https://github.com/zzhang1987/Truncated-Matrix-Power-Iteration-for-Differentiable-DAG-Learning}.
\end{itemize}
\textbf{Platform. }The majority of our experiments are conducted on
a 24-core Intel processor with 4.9 GHz frequency, and an NVIDIA GPU
with 16G of CUDA memory and compute capability of 8.9. The only exception
is the case of CORL on large graphs (Figure~\ref{fig:synthetic}(b))
which requires more than 16Gb of CUDA memory, and therefore these
experiments are conducted on an NVIDIA A100 GPU with 32G of CUDA memory
instead.

Additionally, in this study, we evaluate the performance of various
methods with respect to both the number of steps and runtime, addressing
two independent questions: \textit{``How accurate can a method become
given a fixed number of steps?''} and \textit{``How accurate can
it be within a given runtime?''.} To ensure fair comparisons, we
account for potential biases in measuring performance solely by the
number of DAG evaluations. This is particularly important for methods
like gradient-based approaches (e.g., DAGMA), which may require many
steps but still exhibit low overall runtime.

To address this, we use runtime as a more equitable efficiency metric.
Specifically, we set a high number of steps for all methods (e.g.,
we use $T=800$ iterations instead the default of only $T=5$ for
DAGMA on linear data) and disable early stopping if applicable, to
capture their progression over an extended period of time. We then
truncate the tracking data, which contains performance metrics and
timestamp at every step, either at a fixed number of steps or a specified
runtime, as illustrated in Figure~\ref{fig:synthetic}. This ensures
that the results in the last column of Figure~\ref{fig:synthetic}
are not constrained by the number of steps. For instance, at the 5-minute
mark in the last column of Figure~\ref{fig:synthetic}a, DAGMA completes
approximately 5 million steps compared to only 50,000 steps in the
third column.

\section{Hyperparameters\label{subsec:Hyper-parameters}}

We provide the specific set of hyperparameters for our $\ours$ method
in Table~\ref{tab:Hyperparameters-for-ours}. More details can be
found in our published source code.

\begin{table}[H]
\caption{Hyperparameters for $\protect\ours$. Unless specifically indicated,
the default hyperparameters here are used for all experiments. \label{tab:Hyperparameters-for-ours}}

\centering{}\resizebox{\textwidth}{!}{%
\begin{tabular}{cccc}
\toprule 
\multirow{2}{*}{\textbf{Hyperparameter}} & \multicolumn{3}{c}{\textbf{Experiment}}\tabularnewline
\cmidrule{2-4}
 & \textbf{Linear data} & \textbf{Nonlinear data with GPs} & \textbf{Sachs data}\tabularnewline
\midrule
Normalize data & No & No & Yes\tabularnewline
Scoring function & $S_{\text{BIC-EV}}$ with linear regression & $S_{\text{BIC-NV}}$ with GP regression & $S_{\text{BIC-NV}}$ with GP regression\tabularnewline
Pruning method & Linear pruning & No pruning & CIT pruning\tabularnewline
Batch size $B$ & \multicolumn{3}{c}{64}\tabularnewline
DAG rank $k$ & \multicolumn{3}{c}{8}\tabularnewline
No. training steps $n_{\text{grads}}$ & \multicolumn{3}{c}{10}\tabularnewline
No. preliminary candidates $C$ & \multicolumn{3}{c}{$100,\!000$}\tabularnewline
Optimizer & \multicolumn{3}{c}{Adam}\tabularnewline
Learning rate & \multicolumn{3}{c}{0.1}\tabularnewline
Replay buffer size $n_{\text{replay}}$ & \multicolumn{3}{c}{$1,\!024$}\tabularnewline
No. hidden units & \multicolumn{3}{c}{64}\tabularnewline
Dropout rate & \multicolumn{3}{c}{0.1}\tabularnewline
\bottomrule
\end{tabular}}
\end{table}

For ALIAS and CORL, apart from the recommended default hyperparameters,
for fairness, we also use the same batch size of 64 as ours (except
for CORL which requires the batch size to be at least the number of
nodes, so for 100-node graphs we have to set the batch size to 100),
as well as the same scoring function and pruning method as ours in
each experiment. Regarding COSMO and DAGMA, we use the linear and
nonlinear versions specific to each experiment and the hyperparameters
are set as recommended. We also use the same BIC score and pruner
specific to each experiment to track the progress of all methods in
Figure~\ref{fig:synthetic}.

Regarding the number of evaluations, for all methods, we run more
than needed then cut off at common thresholds.
\begin{rem}
Following the same line as in \citep{Bello_etal_22dagma,Zheng_etal_18DAGs}
and many studies in this field, we intentionally avoid hyperparameter
tuning. This is to prevent injecting spurious information of the dataset
characteristics into the causal discovery results, which is against
the idea of causality. Therefore, we use a fixed set of hyperparameters
for each method in each setting. The hyperparameters of baseline methods
are chosen as recommended in the original manuscripts, while ours
are chosen based on common practice and prior experience. Indeed,
our ablation studies reveal that there are better hyperparameter choices
that can further improve the performance of our method.
\end{rem}

\section{Additional Causal Discovery Settings\label{subsec:Additional-Settings}}

\subsection{Different Sample Sizes}

\begin{figure}[H]
\includegraphics[width=1\textwidth]{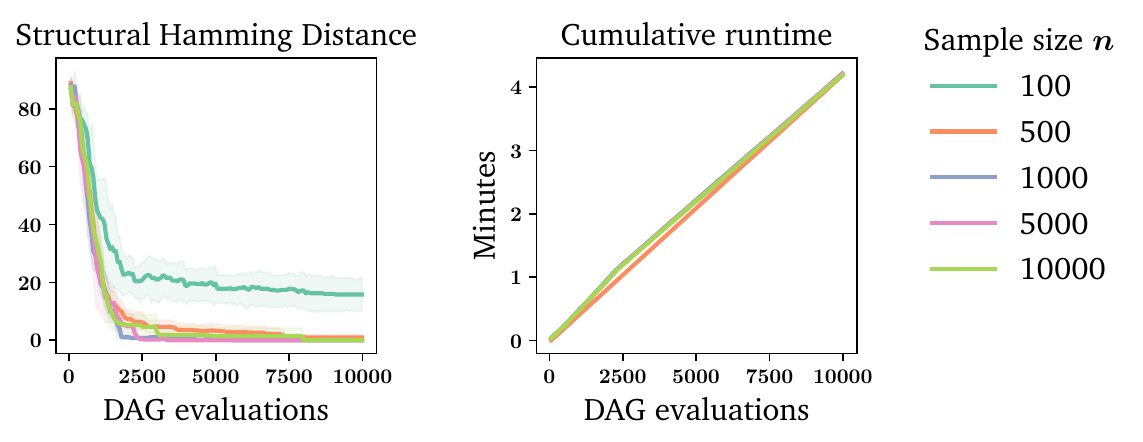}

\caption{\textbf{Causal Discovery Performance with Varying Sample Sizes.} We
apply our $\protect\ours$ method on linear-Gaussian data with 20ER4
graphs. Shaded areas represent 95\% confidence interval over 5 runs.\label{fig:sample-size}}
\end{figure}

In Figure~\ref{fig:sample-size}, we test our method with varying
sample sizes on linear-Gaussian data. The results indicate that our
method is already very accurate with $\text{SHD}\approx0$ at merely
$500$ samples.

\subsubsection{Different Graph Types}

We evaluate our method on different graph types, namely ER and SF,
with varying densities, as shown in Table~\ref{tab:different-graphs}.
Our method perfectly identifies the correct DAG in all cases for both
graph models, even in the dense graphs ER8 and SF8.

\begin{table}[H]

\caption{\textbf{Causal Discovery Performance with Different Graph Types.}
The considered graph models include Erd\H{o}s-R\'{e}nyi \citep{Erdos_Renyi_60Evolution}
and Scale-Free \citep[SF,][]{Barabasi_Albert_99Emergence}, with expected
in-degrees of 2 and 8 corresponding to the case of sparse and dense
structures, respectively. We compare our $\protect\ours$ method with
ALIAS \citep{Duong_etal_24Alias} and DAGMA \citep{Bello_etal_22dagma}
on linear-Gaussian datasets with 20 variables and $10,\!000$ samples.
The numbers are $\text{mean}\pm\text{std}$ over 5 random datasets.
For fairness, all methods are limited to $10,\!000$ score evaluations.\label{tab:different-graphs} }

\resizebox{\textwidth}{!}{

\begin{tabular}{cccccc}
\toprule 
\textbf{Graph} & \textbf{Expected In-degree} & \textbf{Method} & \textbf{SHD $\downarrow$} & \textbf{FDR $\downarrow$} & \textbf{TPR $\uparrow$}\tabularnewline
\midrule 
\multirow{6}{*}{ER} & \multirow{3}{*}{2} & ALIAS & $21.6\pm7.2$ & $0.32\pm0.12$ & $0.68\pm0.08$\tabularnewline
 &  & DAGMA & $25.8\pm7.3$ & $0.28\pm0.13$ & $0.55\pm0.07$\tabularnewline
\cmidrule{3-6}
 &  & $\ours$ (ours) & $\mathbf{\phantom{0}0.0\pm0.0}$ & $\mathbf{0.00\pm0.00}$ & $\mathbf{1.00\pm0.00}$\tabularnewline
\cmidrule{2-6}
 & \multirow{3}{*}{8} & ALIAS & $\phantom{0}84.2\pm7.5$ & $0.15\pm0.03$ & $0.53\pm0.04$\tabularnewline
 &  & DAGMA & $122.4\pm4.7$ & $0.19\pm0.03$ & $0.28\pm0.02$\tabularnewline
\cmidrule{3-6}
 &  & $\ours$ (ours) & $\mathbf{\phantom{00}0.0\pm0.0}$ & $\mathbf{0.00\pm0.00}$ & $\mathbf{1.00\pm0.00}$\tabularnewline
\midrule 
\multirow{6}{*}{SF} & \multirow{3}{*}{2} & ALIAS & $13.6\pm2.7$ & $0.18\pm0.09$ & $0.71\pm0.02$\tabularnewline
 &  & DAGMA & $23.4\pm9.8$ & $0.25\pm0.19$ & $0.53\pm0.13$\tabularnewline
\cmidrule{3-6}
 &  & $\ours$ (ours) & $\mathbf{\phantom{0}0.0\pm0.0}$ & $\mathbf{0.00\pm0.00}$ & $\mathbf{1.00\pm0.00}$\tabularnewline
\cmidrule{2-6}
 & \multirow{3}{*}{8} & ALIAS & $68.8\pm9.0$ & $0.34\pm0.08$ & $0.52\pm0.06$\tabularnewline
 &  & DAGMA & $94.4\pm8.1$ & $0.49\pm0.11$ & $0.2\pm0.06$\tabularnewline
\cmidrule{3-6}
 &  & $\ours$ (ours) & $\mathbf{\phantom{0}0.0\pm0.0}$ & $\mathbf{0.00\pm0.00}$ & $\mathbf{1.00\pm0.00}$\tabularnewline
\bottomrule
\end{tabular}

}
\end{table}

\subsubsection{Different Noise Distributions}

To examine the performance of our method compared with others under
noise misspecification, in Table~\ref{tab:different-noises}, we
evaluate causal discovery performance various noise distributions.
Specifically, we consider linear SCM $x_{i}:=\sum_{j\in\pa_{i}}w_{ji}x_{j}+\varepsilon_{i}$,
where $\varepsilon_{i}$ is drawn from one of the following distributions
\begin{itemize}
\item \textbf{Exponential noise:} $\varepsilon_{i}\sim\text{Exp}\left(1\right),\forall i=1,\ldots,d$,
where 1 is the scale parameter.
\item \textbf{Gaussian noise:} $\varepsilon_{i}\sim\mathcal{N}\left(0,1\right),\forall i=1,\ldots,d$,
where 0 is the mean and 1 is the variance.
\item \textbf{Gumbel noise:} $\varepsilon_{i}\sim\text{Gumbel}\left(0,1\right),\forall i=1,\ldots,d$,
where 0 is the location parameter and 1 is the scale parameter.
\item \textbf{Laplace noise:} $\varepsilon_{i}\sim\text{Laplace}\left(0,1\right),\forall i=1,\ldots,d$,
where 0 is the location parameter and 1 is the scale parameter.
\item \textbf{Uniform noise:} $\varepsilon_{i}\sim\mathcal{U}\left(-1,1\right),\forall i=1,\ldots,d$,
where the minimum value is $-1$ and maximum value is \textbf{$1$}.
\end{itemize}
\begin{rem}
We note that our BIC score is kept unchanged for different noises
to show that our method can work well beyond the assumed Gaussian
noise.
\end{rem}
\begin{table}[H]
\caption{\textbf{Causal Discovery Performance with Different Noise Distributions.}
We consider 5 noise distributions: Exponential, Gaussian, Gumbel,
Laplace, and Uniform. Our $\protect\ours$ method is compared with
ALIAS \citep{Duong_etal_24Alias} and DAGMA \citep{Bello_etal_22dagma}
on linear-Gaussian datasets with 20 variables and $10,\!000$ samples.
The numbers are $\text{mean}\pm\text{std}$ over 5 random datasets.
For fairness, all methods are limited to $20,\!000$ score evaluations.\label{tab:different-noises}}

\resizebox{\textwidth}{!}{
\begin{centering}
\begin{tabular}{ccccc}
\toprule 
\textbf{Noise Type} & \textbf{Method} & \textbf{SHD $\downarrow$} & \textbf{FDR $\downarrow$} & \textbf{TPR $\uparrow$}\tabularnewline
\midrule 
\multirow{3}{*}{Exponential} & ALIAS & $36.2\pm13.3$ & $0.24\pm0.07$ & $0.75\pm0.09$\tabularnewline
 & DAGMA & $56.6\pm6.8$ & $0.22\pm0.10$ & $0.37\pm0.05$\tabularnewline
\cmidrule{2-5}
 & $\ours$ (ours) & $\mathbf{\phantom{0}0.4\pm0.6}$ & $\mathbf{0.00\pm0.01}$ & $\mathbf{0.99\pm0.01}$\tabularnewline
\midrule 
\multirow{3}{*}{Gaussian} & ALIAS & $34.2\pm11.4$ & $0.23\pm0.08$ & $0.77\pm0.05$\tabularnewline
 & DAGMA & $57.2\pm\phantom{0}7.6$ & $0.25\pm0.09$ & $0.39\pm0.09$\tabularnewline
\cmidrule{2-5}
 & $\ours$ (ours) & $\mathbf{\phantom{0}0.0\pm\phantom{0}0.0}$ & $\mathbf{0.00\pm0.00}$ & $\mathbf{1.00\pm0.00}$\tabularnewline
\midrule 
\multirow{3}{*}{Gumbel} & ALIAS & $36.6\pm12.9$ & $0.26\pm0.08$ & $0.77\pm0.06$\tabularnewline
 & DAGMA & $54.8\pm\phantom{0}7.4$ & $0.25\pm0.04$ & $0.44\pm0.08$\tabularnewline
\cmidrule{2-5}
 & $\ours$ (ours) & $\mathbf{\phantom{0}0.2\pm\phantom{0}0.5}$ & $\mathbf{0.00\pm0.01}$ & $\mathbf{1.00\pm0.01}$\tabularnewline
\midrule 
\multirow{3}{*}{Laplace} & ALIAS & $36.4\pm14.2$ & $0.26\pm0.07$ & $0.76\pm0.09$\tabularnewline
 & DAGMA & $56.0\pm\phantom{0}8.8$ & $0.25\pm0.08$ & $0.40\pm0.09$\tabularnewline
\cmidrule{2-5}
 & $\ours$ (ours) & $\mathbf{\phantom{0}0.0\pm\phantom{0}0.0}$ & $\mathbf{0.00\pm0.00}$ & $\mathbf{1.00\pm0.00}$\tabularnewline
\midrule 
\multirow{3}{*}{Uniform} & ALIAS & $34.4\pm13.0$ & $0.24\pm0.08$ & $0.77\pm0.05$\tabularnewline
 & DAGMA & $59.0\pm\phantom{0}7.3$ & $0.26\pm0.08$ & $0.39\pm0.09$\tabularnewline
\cmidrule{2-5}
 & $\ours$ (ours) & $\mathbf{\phantom{0}0.0\pm\phantom{0}0.0}$ & $\mathbf{0.00\pm0.00}$ & $\mathbf{1.00\pm0.00}$\tabularnewline
\bottomrule
\end{tabular}
\par\end{centering}
}
\end{table}

\subsubsection{BGe Score for Markov Equivalence Class Discovery}

Here, we show that our method is not restricted to the BIC score and
can be extended to other scores as well. We consider the Bayesian
Gaussian equivalent \citep[BGe,][]{Geiger_Heckerman_94Learning,Heckerman_etal_1995Learning}
score for learning the Markov Equivalence Class (MEC) of DAGs in linear-Gaussian
settings. The BGe score assigns equal scores to DAGs belonging to
the same MEC, and can be decomposed as the sum of local scores as
follows:

\begin{equation}
S_{\text{BGe}}\left(\mathcal{D},\mathcal{G}\right)=\sum_{i}^{d}\text{LocalBGe}_{i}\left(\pa_{i}^{\mathcal{G}}\right).\label{eq:bge}
\end{equation}

We refer readers to, for example, \citet{Kuipers_etal_14Addendum},
for the specific formula of the BGe score. To adapt our method with
this score, we simply train each dropout network $\text{DropoutNN}_{i}$
to predict $\text{LocalBGe}_{i}$ from $\pa_{i}^{\mathcal{G}}$ and
combine them using Eq.~(\ref{eq:bge}). It can be seen that this
adaptation does not involve changing any other component of our method.

\begin{figure}[H]
\includegraphics[width=1\textwidth]{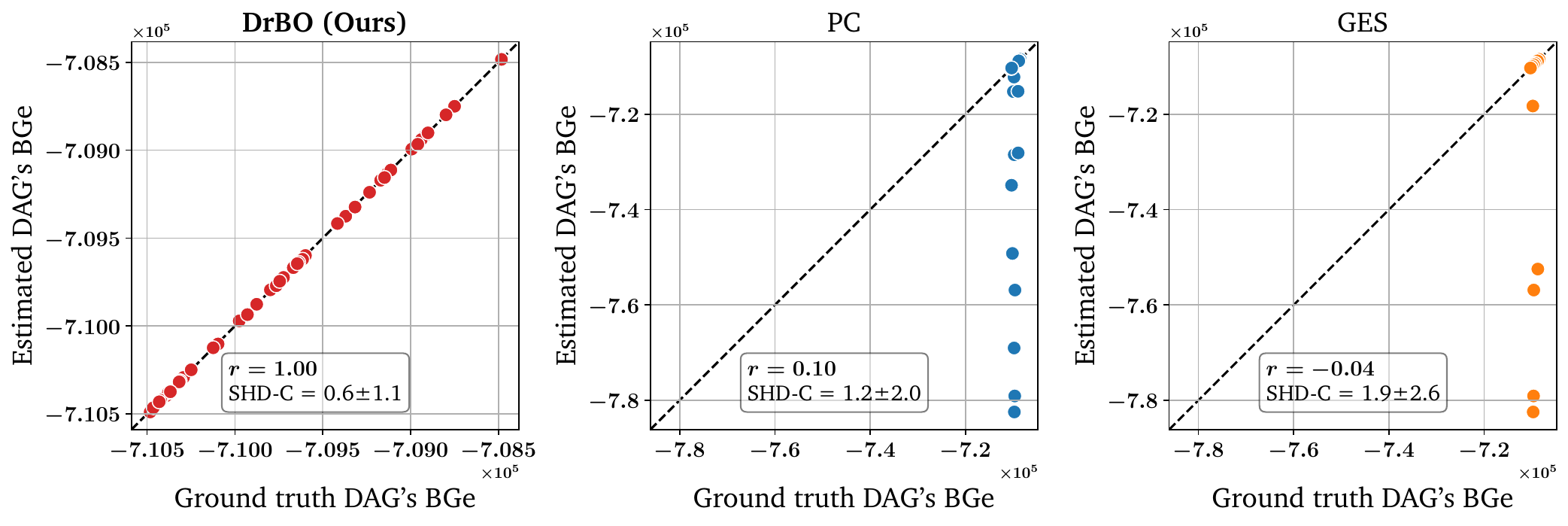}\caption{\textbf{BGe for Markov Equivalence Class Discovery.} We compare the
BGe score of ground truth DAGs and the estimations from $\protect\ours$
with two popular baselines PC \citep{Spirtes_etal_00Causation} and
GES \citep{Chickering_02Optimal}. Each point corresponds to one of
50 random datasets with linear Gaussian data on ER graphs of 5 nodes
and 5 edges on average. The Pearson correlation coefficient $r$ between
the scores of the estimated and ground truth DAGs are included. In
addition, we also report the SHD-C metric, which measures the structural
distance between MECs.\label{fig:bge}}
\end{figure}

We report the results in Figure~\ref{fig:bge}, where we compare
$\ours$ with two popular baselines that are well-known to recover
the MEC, namely PC \citep{Spirtes_etal_00Causation} and GES \citep{Chickering_02Optimal}.\footnote{We use the gCastle library \citep{Zhang_etal_2021Gcastle} for their
implementations, where hyperparameters are left as default.} The results illustrate that our method can find the DAGs with the
highest BGe scores, while the scores from PC' and GES's estimations
are well below those of the ground truths. As a result, our recovered
structures are more accurate than the baselines.

\subsubsection{Discrete Data}

In this section, we show that our method is not limited to continuous
data either. To demonstrate, we consider binary data with logistic
causal models:

\[
x_{i}\sim\text{Bernoulli}\left(f_{i}\left(\mathbf{x}_{\pa_{i}^{\mathcal{G}}}\right)\right).
\]

We adapt our method to this situation by simply changing the BIC score
to take into account logistic models. More particularly, we use the
BIC score for logistic data as in Eq.~(\ref{eq:bic-logistic}), where
the maximum log-likelihood can be decomposed into 
\[
\hat{\mathcal{L}}=\sum_{i=1}^{d}\text{LocalMLL}_{i},
\]

where $\text{LocalMLL}_{i}=\sum_{j=1}^{n}\left(x_{i}^{\left(j\right)}\ln\hat{f_{i}}\left(\mathbf{x}_{\pa_{i}^{\mathcal{G}}}^{\left(j\right)}\right)+\left(1-x_{i}^{\left(j\right)}\right)\ln\left(1-\hat{f_{i}}\left(\mathbf{x}_{\pa_{i}^{\mathcal{G}}}^{\left(j\right)}\right)\right)\right)$,
with $\hat{f_{i}}$ being the maximum-likelihood estimator found via
logistic regression.

Then, we employ our local surrogate models $\text{DropoutNN}_{i}$
to model $\text{LocalMLL}_{i}$, and the acquisition function values
are obtained by summing the local score samples. Again, this adaptation
only involves changing the scoring function and does not modify any
other component of our BO framework.

\begin{figure}[H]
\includegraphics[width=1\textwidth]{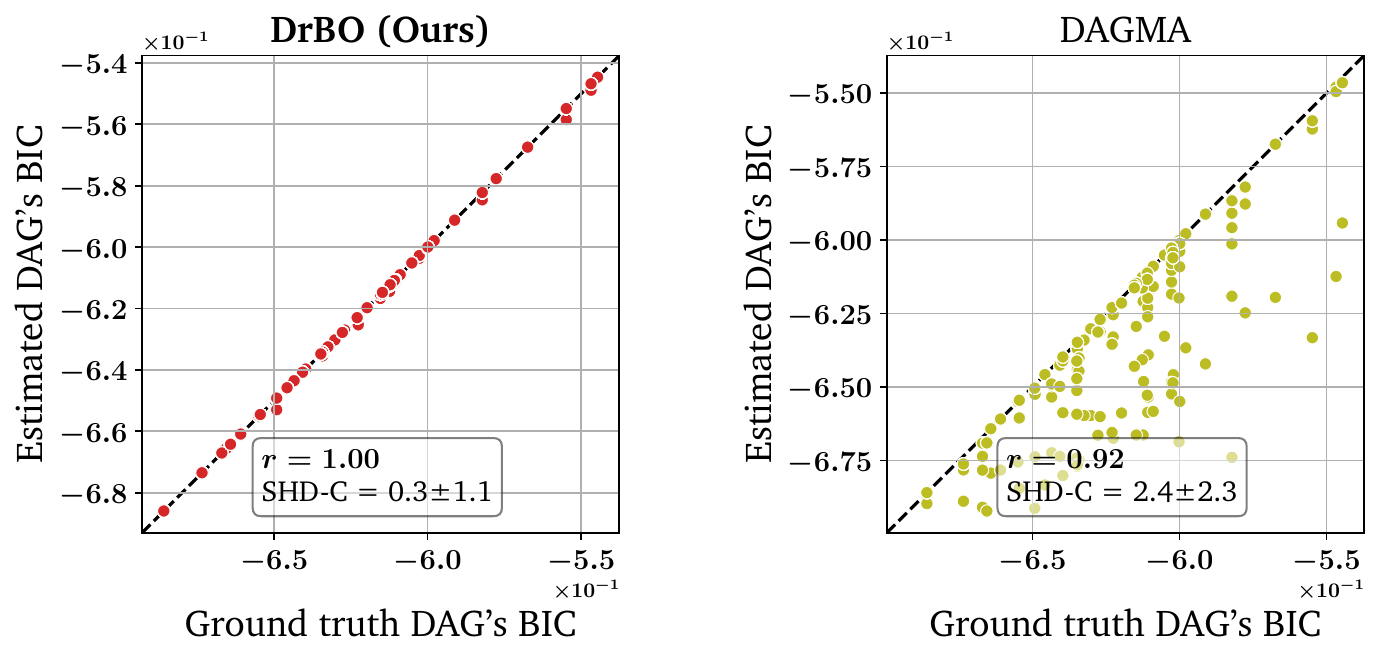}\caption{\textbf{Causal Discovery performance on Binary Data.} We compare our
$\protect\ours$ method using the BIC score for the logistic model
with DAGMA \citep{Bello_etal_22dagma} using the logistic loss. Each
point corresponds to one of 50 random datasets with logistic data
on ER graphs of 5 nodes and 5 edges on average. The Pearson correlation
coefficient $r$ between the scores of the estimated and ground truth
DAGs are included. In addition, we also report the SHD-C metric, which
measures the structural distance between MECs. \label{fig:Discrete-data}}
\end{figure}

We report the causal discovery performance under binary data in Figure~\ref{fig:Discrete-data},
where we also compare $\ours$ with the state-of-the-art DAGMA, which
is now set to use the logistic loss supported. It can be seen that
our method also performs well for binary data, where it consistently
obtains the highest scoring graphs, resulting in a low structural
error. Meanwhile, DAGMA usually finds only sub-optimal solutions which
lead to high structural errors.

\subsubsection{Standardized Data}

As discussed in \citet{Reisach_etall_21Beware}, marginal variances
may contain crucial information about the causal ordering among the
causal variables, and thus revealing the causal DAG by simply sorting
the variables by increasing variances. For this reason, in this section,
we investigate the causal discovery performance of the proposed method
in comparison with the baselines under uninformative marginal variances,
by standardizing the observed data to have zero mean and unit variance
per dimension, before feeding it to causal discovery methods.

\begin{figure}[H]
\resizebox{\columnwidth}{!}{%
\begin{tabular}{>{\centering}p{1\columnwidth}}
\includegraphics[width=1\columnwidth]{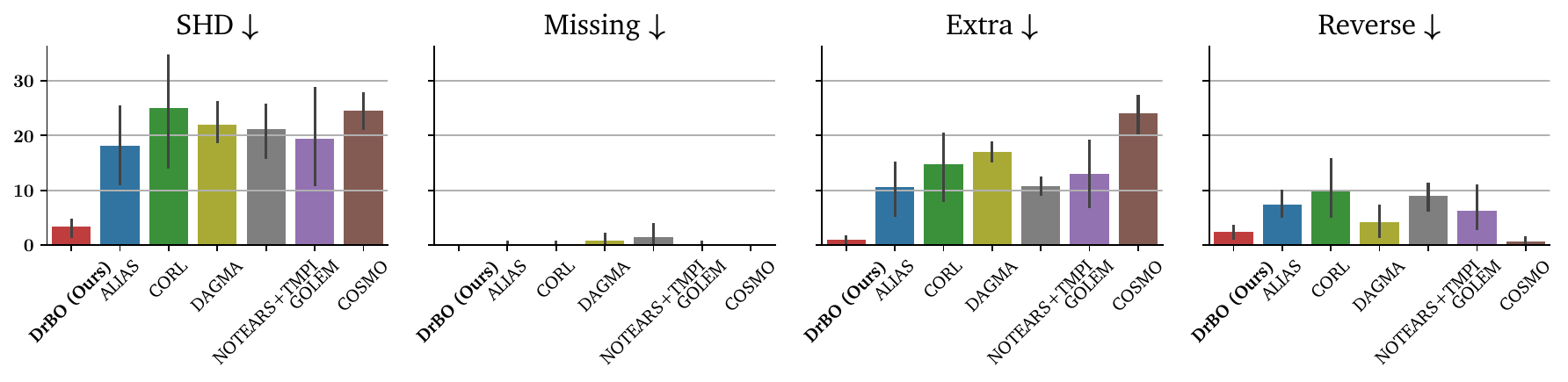}\tabularnewline
\textbf{(a) Linear-Gaussian Data} (DAGs with \textbf{10 nodes} and
\textbf{\ensuremath{\approx}20 edges}). For fairness, all metrics
are calculated at $20,\!000$ evaluations for all methods.\tabularnewline
\includegraphics[width=1\columnwidth]{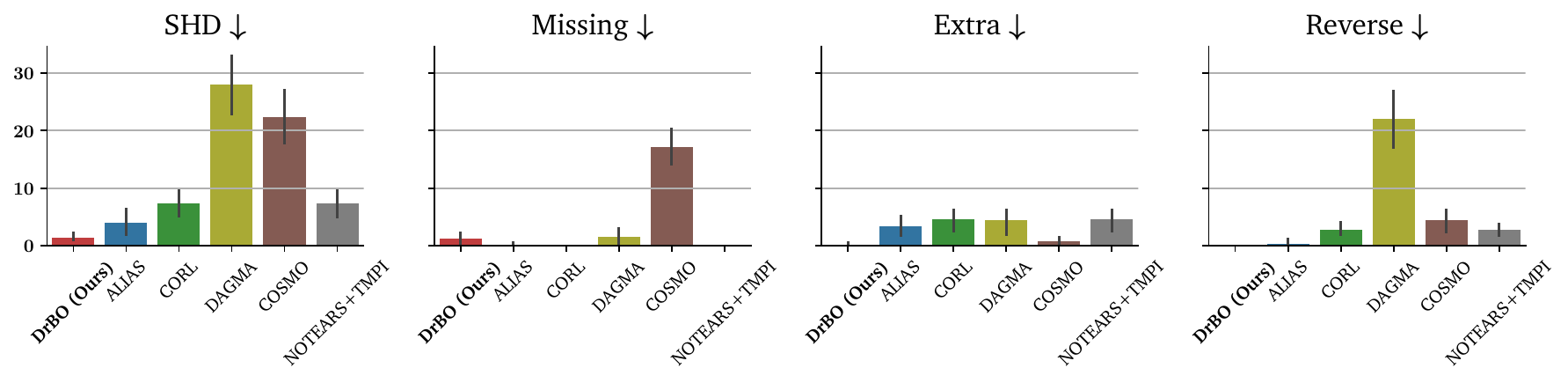}\tabularnewline
\textbf{(b) Nonlinear Data with GPs} (DAGs with \textbf{10 nodes}
and \textbf{\ensuremath{\approx}40 edges}). For fairness, all metrics
are calculated at $20,\!000$ evaluations for all methods.\tabularnewline
\end{tabular}}

\caption{\textbf{Causal Discovery Performance on Standardized Data.} Performance
metrics are Structural Hamming Distance (SHD), number of Missing,
Extra, and Reverse edges. Lower values are more preferable. Error
bars indicate 95\% confidence intervals over 5 simulations.\label{fig:standardized-result}}

\end{figure}

\textbf{Linear-Gaussian data.} Since the noise variances are non-equal
after standardization, we employ the non-equal variance versions of
the methods that support it, including ALIAS, CORL, GOLEM, and $\ours$.
In addition, as the data is standardized, the usual threshold of 0.3
for pruning is no longer appropriate because significant edge weights
may be rescaled to much smaller values after standardization, so in
this experiment, we increase the sample size to $100,\!000$ to reduce
weight estimation variance, and lower the pruning threshold to 0.01.
The results presented in Figure~\ref{fig:standardized-result}(a)
confirm that our method is still robust for standardized data. Overall,
while all methods obtain a non-zero SHD due to the difficulty of standardized
data, which render the equal-variance linear-Gaussian SCM unidentifiable,
our method still outperforms other baselines significantly, where
we achieve an SHD\ensuremath{\approx}3, while the second-best SHD
is nearly 20, highlighting $\ours$'s improved effectiveness over
existing approaches in this intricate scenario. Additionally, even
though the same pruning threshold is used for all methods, our method
barely predicts any extra edge, while other baselines still suffer
from high false discovery rate. Moreover, our method does not predict
too many reverse edges, as opposed to most methods, showing that while
data standardization negatively impacts causal discovery performance
to some extent, the effect on our method is minimal.

\textbf{Nonlinear data with GPs}. Figure~\ref{fig:standardized-result}(b)
presents the results for nonlinear data with GPs with standardization,
showing that our method can achieve a very low SHD and surpasses other
methods considerably. This result is similar to Figure~\ref{fig:synthetic}(c),
where the same datasets employed are not standardized, indicating
that the performance of our method is not affected by data standardization
in this case, which could be potentially thanks to the fact that nonlinear
ANMs remain identifiable after standardization.

\subsubsection{Why low-rank DAG representation tends to perform better\label{subsec:Why-low-rank-DAG}}

In Figure~\ref{fig:Ablating}(a), we have shown that the rank $k$
of our DAG representation plays an important role in the sample-efficiency
of our method, where lower ranks clearly enable reaching the accurate
DAGs earlier than their high-rank counterparts. This is because it
is much more challenging to search in a very high-dimensional space
compared to a lower-dimensional one. Specifically, for higher ranks,
the search space is much larger and sparser than the low-rank ones,
and due to the curse of dimensionality, sampling the same number of
random DAG candidates (step 3 in Algorithm~\ref{alg:ours}) in the
higher-rank search spaces tends to lead to fewer unique candidates
compared with a lower-rank one, reducing the chance to meet the higher-scoring
DAGs earlier.

To empirically verify this, we calculate the number of unique DAGs
among $1,\!000$ random 30-node DAGs generated with different ranks
in Table~\ref{tab:Effect-of-DAG-rank}. It can be seen that, typically,
the lower the rank, the more unique DAGs we can pre-examine for exploration.
For $k=2$, almost every DAG among $1,\!000$ generated DAGs is unique,
and thus our method can reach SHD\ensuremath{\approx}0 very quickly
in Figure~\ref{fig:Ablating}(a), whereas the full-rank representation
is higher-dimensional and can only generate fewer than half the unique
DAGs. In addition, for $k=32\approx d$, where the dimensionality
is highest, we can only generate fewer than $10\%$ of unique DAGs,
explaining why this representation is the least sample-efficient one
in Figure~\ref{fig:Ablating}(a).

\begin{table}[H]
\caption{\textbf{Effect of DAG Rank on Exploration Diversity.} We generate
$1,\!000$ DAGs with $d=30$ nodes using $\mathcal{G}:=\tau\left(\mathbf{z}\right),\mathbf{z}\in\left[-1,1\right]^{d\cdot\left(1+k\right)}$
with different $k$. The numbers are $\text{mean}\pm\text{std}$ over
10 simulations.\label{tab:Effect-of-DAG-rank}}

\resizebox{\columnwidth}{!}{%
\begin{tabular}{ccc}
\toprule 
\textbf{Rank $k$ in Eq.~(\ref{eq:vec2dag-lr})} & \textbf{Number of dimensions} & \textbf{Number of unique 30-node DAGs over $1,\!000$ random DAGs}\tabularnewline
\midrule 
2 & 90 & $926.7\pm\phantom{0}7.0$\tabularnewline
4 & 150 & $779.2\pm12.7$\tabularnewline
8 & 270 & $493.5\pm12.3$\tabularnewline
12 & 390 & $332.4\pm10.8$\tabularnewline
32 & 990 & $\phantom{0}90.7\pm\phantom{0}9.5$\tabularnewline
\midrule 
Full rank \citep[Vec2DAG,][]{Duong_etal_24Alias} & 465 & $421.9\pm13.8$\tabularnewline
\bottomrule
\end{tabular}}

\end{table}

\subsubsection{Large-scale nonlinear data}

\begin{figure}[H]
\begin{centering}
\begin{tabular*}{1\columnwidth}{@{\extracolsep{\fill}}>{\centering}p{1\textwidth}}
\includegraphics[width=1\textwidth]{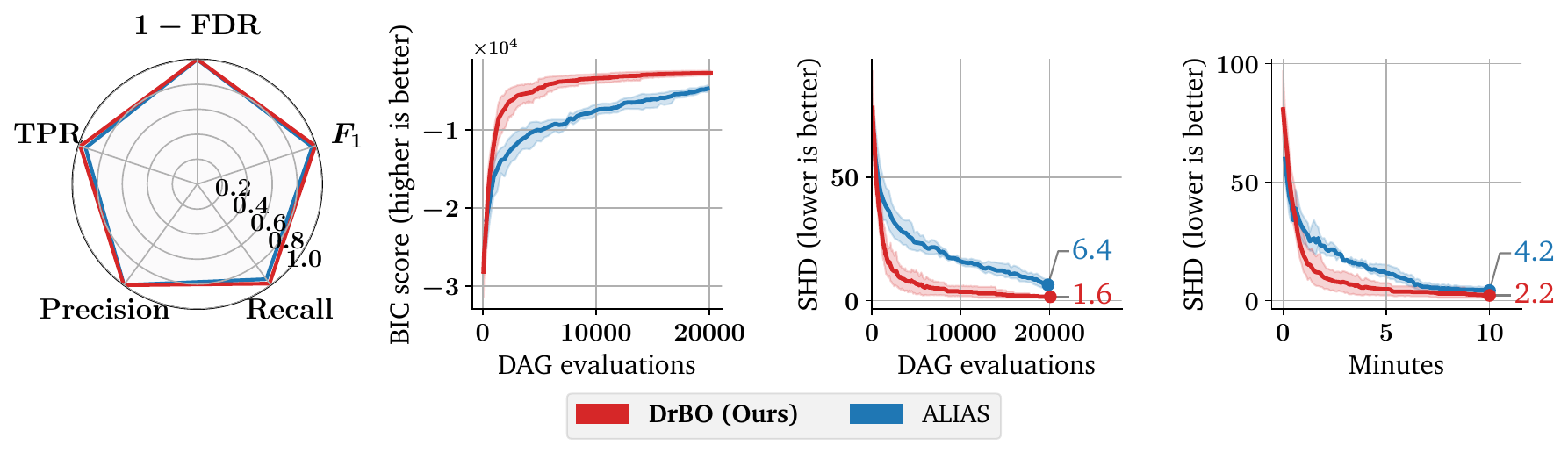}\tabularnewline
\resizebox{0.5\textwidth}{!}{%
\begin{tabular}{ccccc}
\toprule 
\multirow{2}{*}{\textbf{Method}} & \multicolumn{4}{c}{\textbf{Time (minutes) to reach}}\tabularnewline
\cmidrule{2-5}
 & \textbf{SHD=20} & \textbf{SHD=10} & \textbf{SHD=5} & \textbf{SHD=2}\tabularnewline
\midrule 
ALIAS  & $1.8\pm0.6$ & $5.1\pm0.9$ & $8.3\pm2.5$ & $9.3\pm3.1$\tabularnewline
$\ours$ (ours) & $\mathbf{1.1\pm0.5}$ & $\mathbf{2.1\pm1.4}$ & $\mathbf{4.5\pm3.2}$ & $\mathbf{6.8\pm4.2}$\tabularnewline
\bottomrule
\end{tabular}}

\vspace{3mm}
\tabularnewline
\textbf{(a) 50 nodes} and \textbf{\ensuremath{\approx}100 edges}.
For fairness, summary performance metrics (circular plot) are calculated
at $20,\!000$ evaluations for both methods.\tabularnewline
\includegraphics[width=1\textwidth]{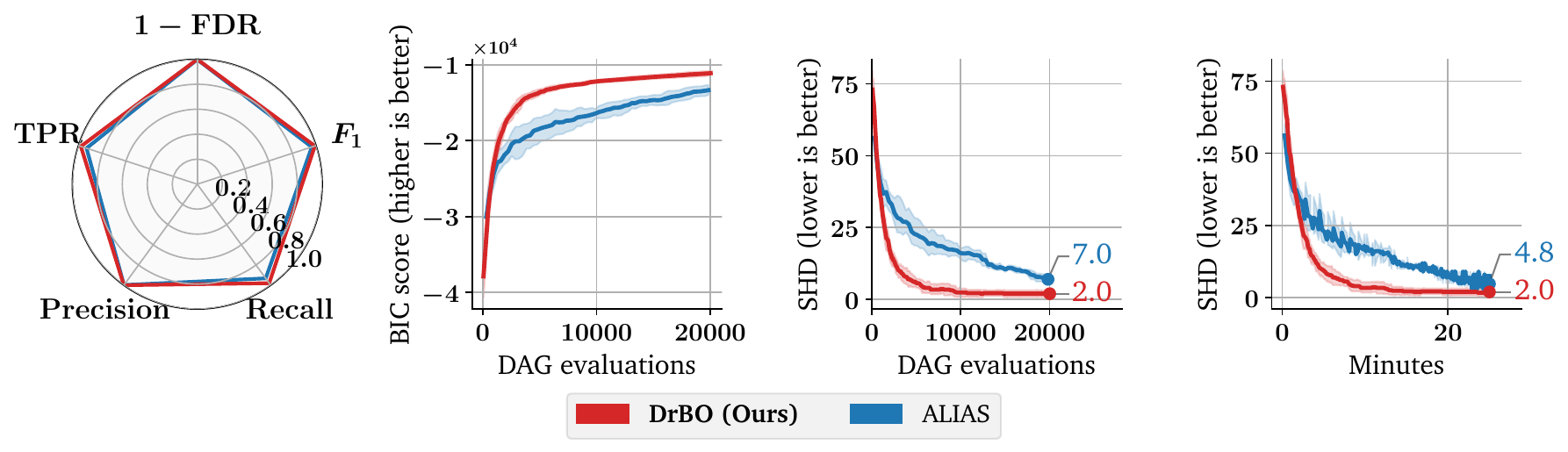}\tabularnewline
\resizebox{0.5\textwidth}{!}{%
\begin{tabular}{ccccc}
\toprule 
\multirow{2}{*}{\textbf{Method}} & \multicolumn{4}{c}{\textbf{Time (minutes) to reach}}\tabularnewline
\cmidrule{2-5}
 & \textbf{SHD=20} & \textbf{SHD=10} & \textbf{SHD=5} & \textbf{SHD=2}\tabularnewline
\midrule 
ALIAS & $6.5\pm2.6$ & $16.3\pm2.5$ & $24.0\pm4.5$ & $33.7\pm6.1$\tabularnewline
$\ours$ (ours) & $\mathbf{2.9\pm0.6}$ & $\mathbf{\phantom{0}5.0\pm1.3}$ & $\mathbf{\phantom{0}8.0\pm3.3}$ & $\mathbf{12.2\pm3.9}$\tabularnewline
\bottomrule
\end{tabular}}\vspace{3mm}
\tabularnewline
\textbf{(b) 100 nodes} and \textbf{\ensuremath{\approx}100 edges}.
For fairness, summary performance metrics (circular plot) are calculated
at $20,\!000$ evaluations for all methods.\tabularnewline
\end{tabular*}
\par\end{centering}
\caption{\textbf{DAG learning results on large-scale nonlinear data.}\label{fig:nonlinear-large}}
\end{figure}

For higher-dimensional data with nonlinearity, following \citet{Zhang_etal_2022Truncated},
we evaluate our method on 50ER2 and 100ER1 graphs with nonlinear SCM
$\mathbf{x}:=\mathbf{B}\cdot\mathrm{cos}\left(\mathbf{x}\right)+{\bf \mathbf{\varepsilon}}$,
where the weights $\mathbf{B}$ are sampled uniformly in $\left[-2,-0.5\right]\cup\left[0.5,2\right]$
and $\varepsilon_{i}\sim\mathcal{N}\left(0,1\right)$. This model
is identifiable according to \citet{Buhlmann_etall_14Cam}. The results
reported in Figure~\ref{fig:nonlinear-large} demonstrate that our
method is also competitive on large-scale nonlinear data, where it
can outperform the baseline in both causal discovery performance and
runtime by a visible margin.

\subsection{Ablation Experiments\label{subsec:Ablation-Experiments}}

\subsubsection{Effect of Dropout Networks\label{subsec:Effect-of-Dropout}}

In Figure~\ref{fig:Ablating}(b), we compare our dropout networks
as the surrogate model with exact GPs and approximate GPs. Approximate
GPs learn a set of pseudo data points called inducing points and conduct
inference via these points instead of the real data \citep{Hensman_etal_15Scalable}.
Here we use a small number of 100 inducing points for Approximate
GPs, to see they can scale well with few inducing points. Due to the
limited scalability and intensive memory requirement for GPs, we can
only use $C=10,\!000$ preliminary candidates on which we sample from
the posteriors, while our default hyperparameter is $C=100,\!000$
using dropout networks.

\subsubsection{Effect of Continual Training\label{subsec:Effect-of-Continual}}

In Figure~\ref{fig:Ablating}(d), we compare the continual training
approach with fully retraining using all data. For continual learning,
we use default hyperparameters $n_{\text{replay}}=1,\!024$, $B=64$,
and $n_{\text{grads}}=10$, meaning for each BO iteration, we perform
10 gradients update, each update is calculated from $1,\!024+64=1,\!088$
datapoints. To ensure fairness, we also use $n_{\text{grads}}=10$
epochs and a mini-batch size of $1,\!088$ for the full retraining
approach.

\subsubsection{Effect of Evaluation Batch Size ($B$)}

\begin{figure}[H]
\centering{}\includegraphics[width=0.8\textwidth]{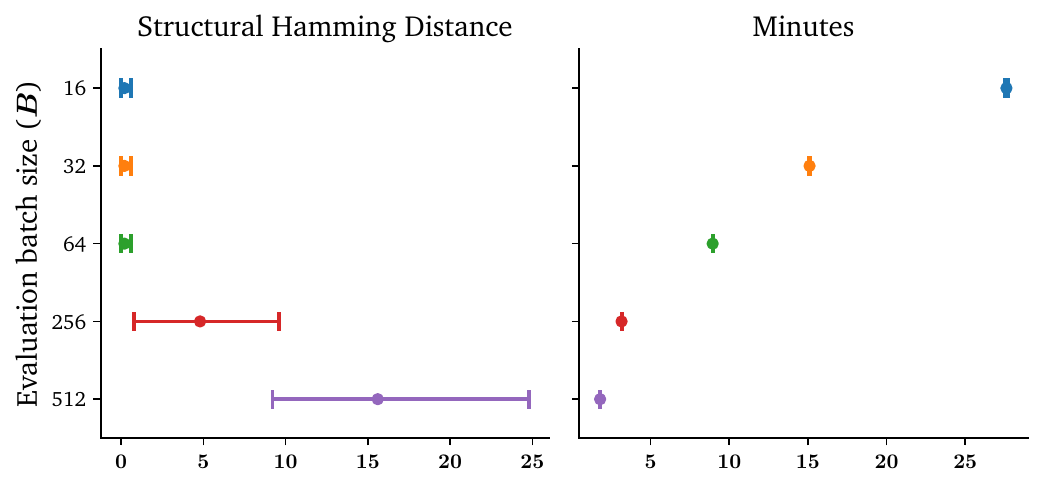}\caption{\textbf{Effect of Evaluation Batch Size $B$.} We evaluate our method
on linear-Gaussian data with 20ER4 graphs and $1,\!000$ observations.
Error bars indicate 95\% confidence intervals over 5 runs. The number
of evaluations is limited to $20,\!000$.\label{fig:ablation-batchsize}}
\end{figure}

In Figure~\ref{fig:ablation-batchsize}, we show the influence of
the evaluation batch size $B$ onto the performance and scalability
of our method. Overall, it is clear that smaller batch sizes lead
to better SHD but much worse runtime since the surrogate model is
updated more frequently, and vice versa. However, $B=64$ seems to
achieve balance, where it enables $\mathrm{SHD}\approx0$ and lower
runtime than smaller batch sizes.

\subsubsection{Effect of Number of Preliminary Candidates ($C$)}

\begin{figure}[H]
\centering{}\includegraphics[width=0.8\textwidth]{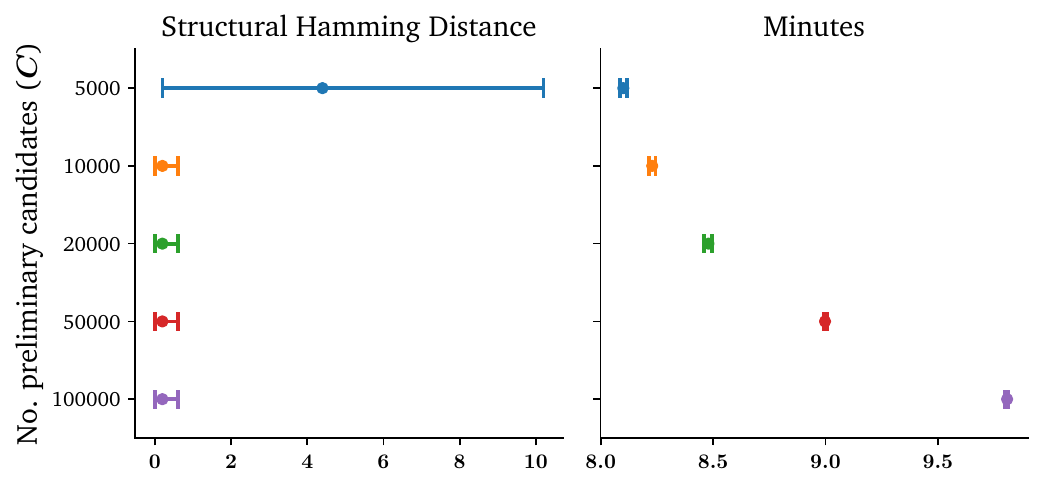}\caption{\textbf{Effect of Number of Preliminary Candidates $C$.} We evaluate
our method on linear-Gaussian data with 20ER4 graphs and $1,\!000$
observations. Error bars indicate 95\% confidence intervals over 5
runs. The number of evaluations is limited to $20,\!000$.\label{fig:ablation-ncands}}
\end{figure}

We show the variation our $\ours$'s performance w.r.t. different
numbers of preliminary candidates $C$ in Figure~\ref{fig:ablation-ncands},
showing that the best performance and runtime can be achieved at $10,\!000$
candidates, and even with 10x more candidates, the runtime of our
method increases only slightly. 

\subsubsection{Effect of Number of Training Steps per BO Iteration ($n_{\text{grads}}$)}

\begin{figure}[H]
\centering{}\includegraphics[width=0.8\textwidth]{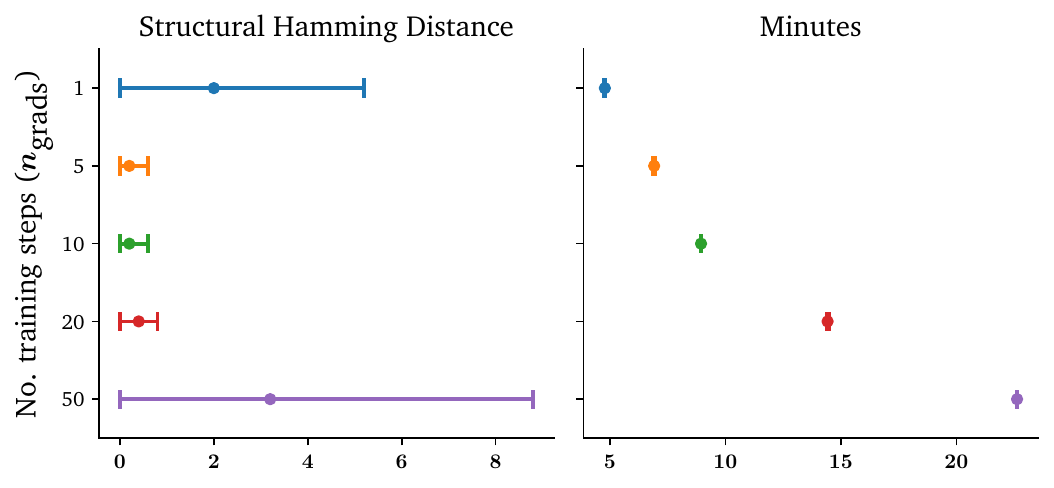}\caption{\textbf{Effect of Number of Training Steps per BO Iteration $n_{\text{grads}}$.}
We evaluate our method on linear-Gaussian data with 20ER4 graphs and
$1,\!000$ observations. Error bars indicate 95\% confidence intervals
over 5 runs. The number of evaluations is limited to $20,\!000$.\label{fig:ablation-ngrads}}
\end{figure}

We study the effect of the number of gradient steps in each BO iteration
($n_{\text{grads}}$) in Figure~\ref{fig:ablation-ngrads}. In general,
small values may lead to underfitting and large values may be prone
to overfitting, so values in the middle are better for this hyperparameter.

\subsubsection{Effect of Replay Buffer Size ($n_{\text{replay}}$)}

\begin{figure}[H]
\centering{}\includegraphics[width=0.8\textwidth]{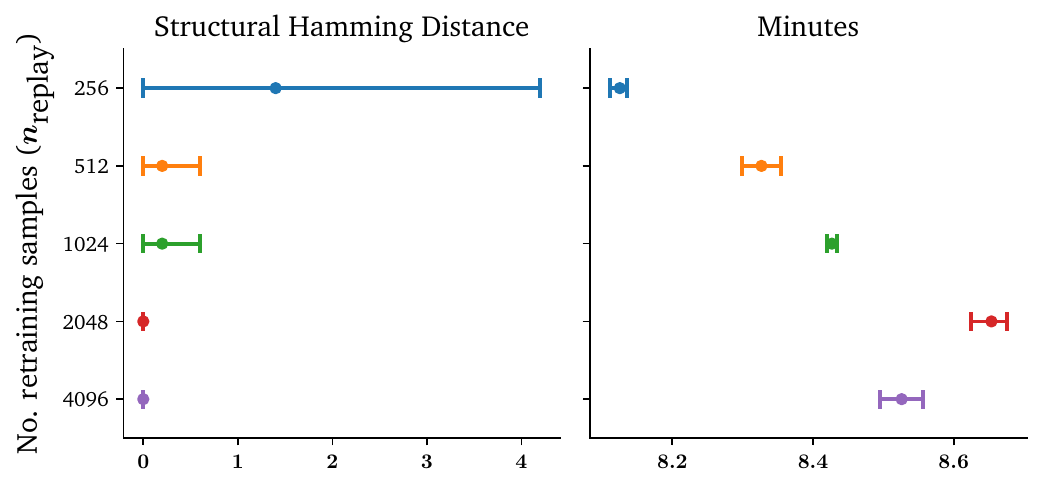}\caption{\textbf{Effect of Replay Buffer Size $n_{\text{replay}}$.} We evaluate
our method on linear-Gaussian data with 20ER4 graphs and $1,\!000$
observations. Error bars indicate 95\% confidence intervals over 5
runs. The number of evaluations is limited to $20,\!000$.\label{fig:ablation-buffersize}}
\end{figure}

In Figure~\ref{fig:ablation-buffersize}, we show that higher values
for the replay buffer size significantly reduces SHD but does not
considerately influence the runtime.

\subsubsection{Effect of Learning Rate}

\begin{figure}[H]
\centering{}\includegraphics[width=0.8\textwidth]{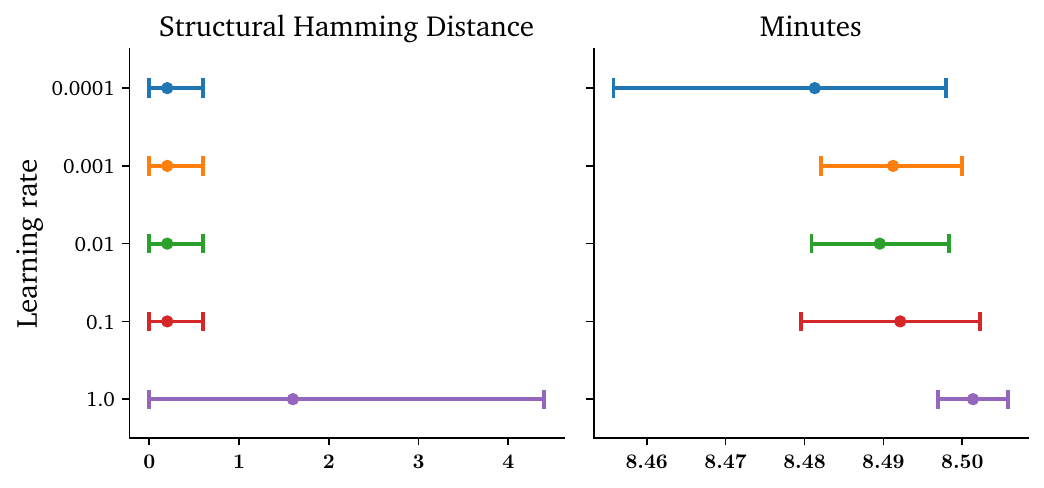}\caption{\textbf{Effect of Learning Rate.} We evaluate our method on linear-Gaussian
data with 20ER4 graphs and $1,\!000$ observations. Error bars indicate
95\% confidence intervals over 5 runs. The number of evaluations is
limited to $20,\!000$.\label{fig:ablation-lr}}
\end{figure}

Figure~\ref{fig:ablation-lr} depicts that the learning rate has
a weak effect on the performance and scalability of our method, where
any value below 1 can achieve the same level of SHD and runtime. The
SHD only becomes large for a high learning rate of 1.

\subsubsection{Effect of Number of Hidden Units}

\begin{figure}[H]
\centering{}\includegraphics[width=0.8\textwidth]{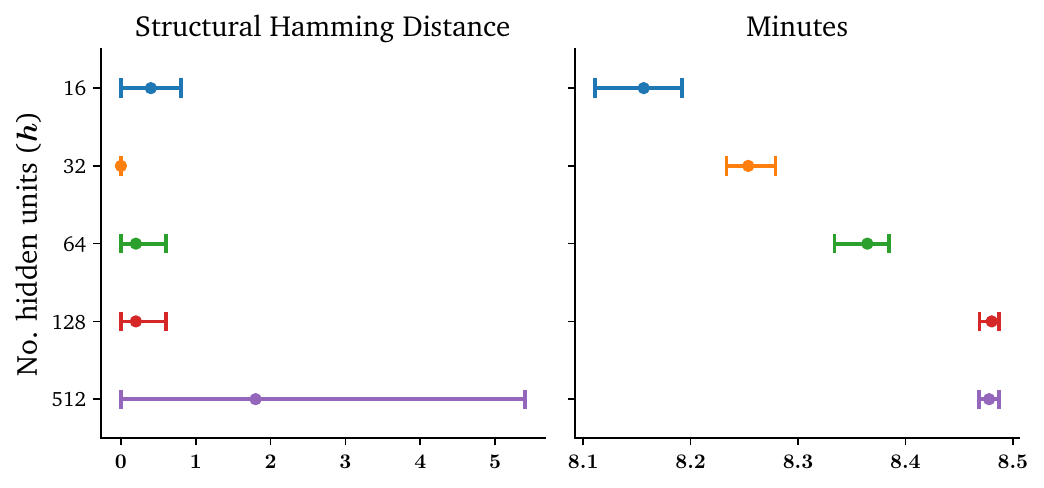}\caption{\textbf{Effect of Number of Hidden Units $h$.} We evaluate our method
on linear-Gaussian data with 20ER4 graphs and $1,\!000$ observations.
Error bars indicate 95\% confidence intervals over 5 runs. The number
of evaluations is limited to $20,\!000$.\label{fig:ablation-hiddensize}}
\end{figure}

We present in Figure~\ref{fig:ablation-hiddensize} that the number
of hidden units in our dropout networks also has a visible effect
on our method's performance, but not much on the runtime. Specifically,
a moderate value of 32 achieves a vanishing SHD with the equivalent
runtime as others. Meanwhile, to many hidden units may challenge the
training process so performance may drop.

\subsubsection{Effect of Number of Dropout Rate}

\begin{figure}[H]
\centering{}\includegraphics[width=0.8\textwidth]{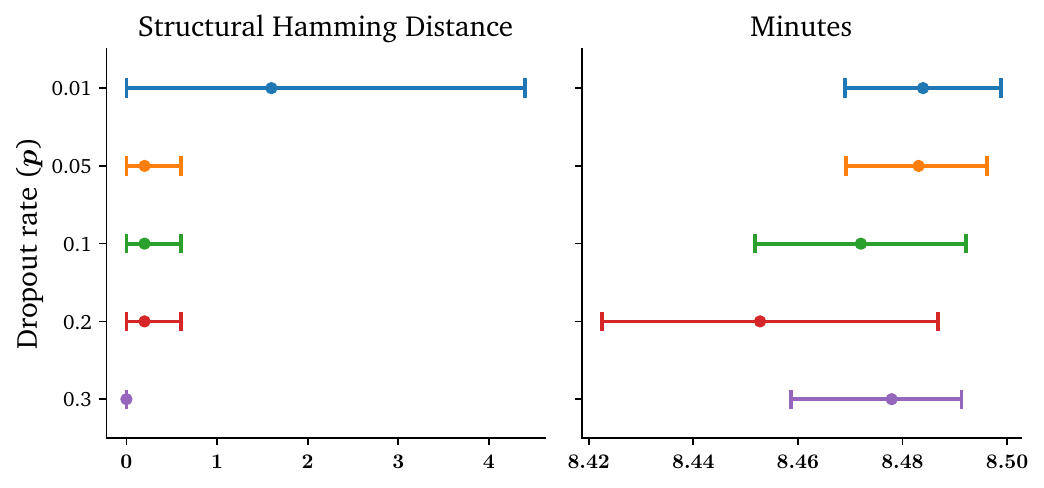}\caption{\textbf{Effect of Dropout Rate $p$.} We evaluate our method on linear-Gaussian
data with 20ER4 graphs and $1,\!000$ observations. Error bars indicate
95\% confidence intervals over 5 runs. The number of evaluations is
limited to $20,\!000$.\label{fig:ablation-dropout}}
\end{figure}

Figure~\ref{fig:ablation-dropout} suggests that the performance
of our method improves with higher dropout rates, while the runtime
does not vary significantly.

\section{Additional Baselines\label{subsec:additional-baselines}}

Apart from the score-based competitors compared so far, here we also
consider additional baselines for comparison, such as conventional
methods PC \citep{Spirtes_etal_00Causation} and GES \citep{Chickering_02Optimal}.
In addition, ordering-based methods have also been gaining popularity
\citep{Reisach_etall_21Beware,Rolland_etal_22Score,Sanchez_etal_22Diffusion,Montagna_23Scalable}.
These methods sidestep the acyclicity issue by first learning a topological
ordering of the causal DAG, then pruning the fully-connected DAG induced
by the ordering to obtain the causal structure.

In Table~\ref{tab:Comparison-with-Ordering-based}, we provide a
comprehensive comparison between our method and conventional methods
PC and GES, as well as popular ordering-based method, including sortnregress
\citep{Reisach_etall_21Beware}, SCORE \citep{Rolland_etal_22Score},
and DiffAN \citep{Sanchez_etal_22Diffusion}, on linear, nonlinear,
as well as real data. The empirical evaluations indicate that our
method is also able to surpass these methods in all metrics and settings.

\begin{table}[H]
\caption{\textbf{Comparison with Additional Baselines.} We compare our method
with conventional methods PC \citep{Spirtes_etal_00Causation} and
GES \citep{Chickering_02Optimal}, as well as popular ordering-based
algorithms sortnregress \citep{Reisach_etall_21Beware}, SCORE \citep{Rolland_etal_22Score},
and DiffAN \citep{Sanchez_etal_22Diffusion}. The orderings produced
by those methods are first transformed into respective fully connected
DAGs. Then, for fairness, all DAGs, including ours, are pruned using
the weight matrix thresholded at $0.3$ for linear data, CAM pruning
\citep{Buhlmann_etall_14Cam} for nonlinear data, and KCIT \citep{Zhang_etal_11Kernel}
for the Sachs dataset. The figures are $\text{mean}\pm\text{std}$
over 5 datasets, except for the Sachs dataset. Note that $\protect\ours$'s
result in Figure~\ref{fig:synthetic}(c) with $\text{SHD}\approx0.4$
is obtained without pruning, while using CAM pruning leads to a few
missing edges as shown below.\label{tab:Comparison-with-Ordering-based}}

\resizebox{\textwidth}{!}{

\begin{tabular}{>{\raggedright}p{0.2\columnwidth}lcccc}
\toprule 
\textbf{Dataset} & \textbf{Method} & \textbf{SHD $\downarrow$} & \textbf{FDR $\downarrow$} & \textbf{TPR $\uparrow$} & \textbf{$F_{1}$ $\uparrow$}\tabularnewline
\midrule 
\multirow{6}{0.2\columnwidth}{Linear Data (10ER2 graphs, $1,\!000$ samples)} & PC & $10.8\pm5.1$  & $0.29\pm0.1$ & $0.49\pm0.2$ & $0.56\pm0.2$\tabularnewline
 & GES & $11.6\pm6.4$ & $0.40\pm0.2$ & $0.61\pm0.2$ & $0.58\pm0.2$\tabularnewline
 & sortnregress & $\phantom{0}2.0\pm2.6$ & $0.10\pm0.1$ & $0.94\pm0.1$ & $0.92\pm0.1$\tabularnewline
 & SCORE & $\mathbf{\phantom{0}0.0\pm0.0}$ & $\mathbf{\phantom{0}0.0\pm0.0}$ & $\mathbf{1.00\pm0.0}$ & $\mathbf{1.00\pm0.0}$\tabularnewline
 & DiffAN & $16.6\pm4.9$ & $0.54\pm0.1$ & $0.49\pm0.1$ & $0.48\pm0.1$\tabularnewline
\cmidrule{2-6}
 & $\ours$ (ours) & $\mathbf{\phantom{0}0.0\pm0.0}$ & $\mathbf{\phantom{0}0.0\pm0.0}$ & $\mathbf{1.00\pm0.0}$ & $\mathbf{1.00\pm0.0}$\tabularnewline
\midrule 
\multirow{6}{0.2\columnwidth}{Linear Data (30ER8 graphs, $1,\!000$ samples)} & PC & $227.0\pm8.6$ & $0.44\pm0.0$ & $0.10\pm0.0$ & $0.17\pm0.0$\tabularnewline
 & GES & \multicolumn{4}{c}{Failed to halt}\tabularnewline
 & sortnregress & $101.4\pm21.8$ & $0.24\pm0.0$ & $0.80\pm0.1$ & $0.78\pm0.0$\tabularnewline
 & SCORE & $247.0\pm11.8$ & $0.76\pm0.1$ & $0.05\pm0.0$ & $0.08\pm0.1$\tabularnewline
 & DiffAN & $236.0\pm\phantom{0}3.7$ & $0.58\pm0.1$ & $0.16\pm0.1$ & $0.23\pm0.1$\tabularnewline
\cmidrule{2-6}
 & $\ours$ (ours) & $\mathbf{\phantom{00}1.6\pm\phantom{0}1.5}$ & $\mathbf{0.00\pm0.0}$ & $\mathbf{0.99\pm0.0}$ & $\mathbf{1.0\pm0.0}$\tabularnewline
\cmidrule{2-6}
\multirow{6}{0.2\columnwidth}{Nonlinear Data (10ER4 graphs, $1,\!000$ samples)} & PC & $32.2\pm2.2$ & $0.50\pm0.2$ & $0.21\pm0.1$ & $0.29\pm0.1$\tabularnewline
 & GES & $29.6\pm5.6$ & $0.43\pm0.2$ & $0.30\pm0.1$ & $0.38\pm0.1$\tabularnewline
 & sortnregress & $\phantom{0}27.4\pm3.3$ & $0.42\pm0.1$ & $0.37\pm0.1$ & $0.45\pm0.1$\tabularnewline
 & SCORE & $\phantom{00}9.4\pm4.6$ & $0.11\pm0.1$ & $0.83\pm0.1$ & $0.86\pm0.1$\tabularnewline
 & DiffAN & $\phantom{0}18.6\pm3.8$ & $0.34\pm0.1$ & $0.61\pm0.1$ & $0.63\pm0.1$\tabularnewline
\cmidrule{2-6}
 & $\ours$ (ours) & $\mathbf{\phantom{00}4.2\pm1.3}$ & $\mathbf{0.00\pm0.0}$ & $\mathbf{0.90\pm0.0}$ & $\mathbf{0.95\pm0.0}$\tabularnewline
\midrule
\multirow{6}{0.2\columnwidth}{\citet{Sachs_etall_05Causal}, (11 nodes, 17 edges, 853 samples)} & PC & 11 & 0.25 & 0.35 & 0.39\tabularnewline
 & GES & 11 & 0.25 & 0.35 & 0.39\tabularnewline
 & sortnregress & $13$ & $0.44$ & $0.29$ & $0.38$\tabularnewline
 & SCORE & $12$ & $0.33$ & $0.35$ & $0.46$\tabularnewline
 & DiffAN & $16$ & $0.75$ & $0.12$ & $0.16$\tabularnewline
\cmidrule{2-6}
 & $\ours$ (ours) & $\mathbf{9}$ & $\mathbf{0.11}$ & $\mathbf{0.47}$ & $\mathbf{0.62}$\tabularnewline
\bottomrule
\end{tabular}

}
\end{table}

\section{Time Comparisons\label{subsec:Time-Comparisons}}

\begin{table}[H]
\caption{Runtime comparison on 30ER8 linear-Gaussian data (experiment in Figure~\ref{fig:synthetic}(a)).}

\centering{}\resizebox{\textwidth}{!}{%
\begin{tabular}{ccccccc}
\toprule 
\multirow{3}{*}{Method} & \multicolumn{6}{c}{\textbf{Max number of evaluations}}\tabularnewline
\cmidrule{2-7}
 & \multicolumn{2}{c}{$10,\!000$} & \multicolumn{2}{c}{$20,\!000$} & \multicolumn{2}{c}{$50,\!000$}\tabularnewline
\cmidrule{2-7}
 & \textbf{SHD} & \textbf{Runtime (mins)} & \textbf{SHD} & \textbf{Runtime (mins)} & \textbf{SHD} & \textbf{Runtime (mins)}\tabularnewline
\midrule 
ALIAS & $208.8\pm13.1$ & $0.5\pm0.0$ & $165.0\pm18.0$ & $1.1\pm0.0$ & $105.2\pm2.9$ & $\phantom{0}2.4\pm0.0$\tabularnewline
CORL & $169.0\pm17.1$ & $1.1\pm0.0$ & $162.0\pm13.4$ & $2.1\pm0.0$ & $145.4\pm29.3$ & $\phantom{0}5.2\pm0.1$\tabularnewline
COSMO & $213.6\pm17.4$ & $0.6\pm0.0$ & $203.4\pm21.0$ & $1.1\pm0.0$ & $195.8\pm10.8$ & $\phantom{0}2.8\pm0.0$\tabularnewline
DAGMA & $222.2\pm11.2$ & $\mathbf{0.0\pm0.0}$ & $218.8\pm10.6$ & $\mathbf{0.0\pm0.0}$ & $181.6\pm22.4$ & $\mathbf{\phantom{0}0.0\pm0.0}$\tabularnewline
\midrule 
$\ours$ (ours) & $\mathbf{\phantom{00}2.0\pm\phantom{0}1.2}$ & $4.6\pm0.0$ & $\mathbf{\phantom{00}2.0\pm\phantom{0}1.4}$ & $9.2\pm0.0$ & $\mathbf{\phantom{00}1.6\pm\phantom{0}1.5}$ & $22.9\pm0.1$\tabularnewline
\bottomrule
\end{tabular}}
\end{table}

\begin{table}[H]
\caption{Runtime comparison on 100ER2 linear-Gaussian data (experiment in Figure~\ref{fig:synthetic}(b)).}

\centering{}\resizebox{\textwidth}{!}{%
\begin{tabular}{ccccccc}
\toprule 
\multirow{3}{*}{Method} & \multicolumn{6}{c}{\textbf{Max number of evaluations}}\tabularnewline
\cmidrule{2-7}
 & \multicolumn{2}{c}{$10,\!000$} & \multicolumn{2}{c}{$20,\!000$} & \multicolumn{2}{c}{$50,\!000$}\tabularnewline
\cmidrule{2-7}
 & \textbf{SHD} & \textbf{Runtime (mins)} & \textbf{SHD} & \textbf{Runtime (mins)} & \textbf{SHD} & \textbf{Runtime (mins)}\tabularnewline
\midrule 
ALIAS & $230.4\pm32.7$ & $\phantom{0}2.7\pm0.3$ & $136.8\pm30.2$ & $\phantom{0}5.5\pm0.3$ & $\phantom{0}32.0\pm16.7$ & $13.5\pm0.6$\tabularnewline
CORL & $148.0\pm42.9$ & $11.1\pm0.1$ & $120.2\pm18.1$ & $22.0\pm0.2$ & $\phantom{0}81.0\pm19.9$ & $54.2\pm0.5$\tabularnewline
COSMO & $111.0\pm10.9$ & $\phantom{0}0.8\pm0.0$ & $111.8\pm12.2$ & $\phantom{0}1.5\pm0.0$ & $112.6\pm13.4$ & $\phantom{0}3.7\pm0.0$\tabularnewline
DAGMA & $124.8\pm17.4$ & $\phantom{0}0.0\pm0.0$ & $\phantom{0}93.6\pm16.1$ & $\phantom{0}0.1\pm0.0$ & $\phantom{00}6.6\pm\phantom{0}4.0$ & $\phantom{0}0.3\pm0.0$\tabularnewline
\midrule 
$\ours$ (ours) & $\mathbf{\phantom{0}29.2\pm16.7}$ & $12.7\pm0.1$ & $\mathbf{\phantom{00}3.4\pm\phantom{0}4.3}$ & $25.4\pm0.3$ & $\mathbf{\phantom{00}1.4\pm\phantom{0}1.1}$ & $62.7\pm0.8$\tabularnewline
\bottomrule
\end{tabular}}
\end{table}

\begin{table}[H]
\caption{Runtime comparison on 10ER4 nonlinear data with Gaussian processes
(experiment in Figure~\ref{fig:synthetic}(c)).}

\centering{}\resizebox{\textwidth}{!}{%
\begin{tabular}{ccccccc}
\toprule 
\multirow{3}{*}{\textbf{Method}} & \multicolumn{6}{c}{\textbf{Max number of evaluations}}\tabularnewline
\cmidrule{2-7}
 & \multicolumn{2}{c}{$1,\!000$} & \multicolumn{2}{c}{$2,\!000$} & \multicolumn{2}{c}{$20,\!000$}\tabularnewline
\cmidrule{2-7}
 & \textbf{SHD} & \textbf{Runtime (mins)} & \textbf{SHD} & \textbf{Runtime (mins)} & \textbf{SHD} & \textbf{Runtime (mins)}\tabularnewline
\midrule 
ALIAS & $15.8\pm4.4$ & $4.0\pm0.2$ & $12.6\pm4.2$ & $5.8\pm0.2$ & $4.0\pm2.3$ & $8.0\pm0.3$\tabularnewline
CORL & $10.4\pm3.4$ & $20.9\pm3.0$ & $9.2\pm2.3$ & $26.5\pm3.1$ & $8.4\pm3.0$ & $29.9\pm2.4$\tabularnewline
COSMO & $34.8\pm2.6$ & $0.3\pm0.1$ & $33.0\pm3.3$ & $0.4\pm0.1$ & $25.6\pm4.1$ & $1.8\pm0.1$\tabularnewline
DAGMA & $40.4\pm2.3$ & $0.0\pm0.0$ & $38.4\pm2.7$ & $0.1\pm0.0$ & $34.2\pm3.5$ & $0.8\pm0.1$\tabularnewline
\midrule 
$\ours$ (ours) & $\mathbf{\phantom{}2.2\pm1.6}$ & $3.7\pm0.8$ & $\mathbf{0.4\pm0.5}$ & $3.9\pm0.9$ & $\mathbf{0.4\pm0.5}$ & $6.1\pm1.0$\tabularnewline
\bottomrule
\end{tabular}}
\end{table}

\end{document}